\documentclass[conference]{IEEEtran}
\usepackage{times}

\usepackage[numbers]{natbib}
\usepackage{multicol}
\usepackage[bookmarks=true]{hyperref}

\usepackage{bbm}
\usepackage{amsmath}
\usepackage{amsfonts}
\usepackage{amssymb}
\usepackage{graphicx}
\usepackage{float}
\usepackage[ruled]{algorithm2e}
\usepackage{etoolbox}
\usepackage{cuted}
\usepackage{capt-of}

\DeclareMathOperator*{\argmin}{arg\,min}

\begin{document}

% \title{Learning Annotation-Free Perception and Planning for Off-Road Driving}

%flashier option
\title{TravSUITE: Traversability via \underline{S}elf-Supervised, \underline{U}ncertainty-Aware \underline{I}RL and \underline{T}errain \underline{E}stimation}

% You will get a Paper-ID when submitting a pdf file to the conference system
% \author{Author Names Omitted for Anonymous Review. Paper-ID 705}

%authors are prob Sam, Matthew, David, Wenshan, Amir, Basti and Anthony if stuff works well

% \author{\authorblockN{Michael Shell}
% \authorblockA{School of Electrical and\\Computer Engineering\\
% Georgia Institute of Technology\\
% Atlanta, Georgia 30332--0250\\
% Email: mshell@ece.gatech.edu}
% \and
% \authorblockN{Homer Simpson}
% \authorblockA{Twentieth Century Fox\\
% Email: homer@thesimpsons.com}
% \and
% \authorblockN{James Kirk\\ and Montgomery Scott}
% \authorblockA{Starfleet Academy\\
% San Francisco, California 96678-2391\\
% Telephone: (800) 555--1212\\

% avoiding spaces at the end of the author lines is not a problem with
% conference papers because we don't use \thanks or \IEEEmembership

\author{\authorblockN{Samuel Triest\authorrefmark{1},
Amirreza Shaban\authorrefmark{2},
David D. Fan\authorrefmark{2}, 
Wenshan Wang\authorrefmark{1} and
Sebastian Scherer\authorrefmark{1}}
\authorblockA{\authorrefmark{1}Robotics Institute, Carnegie Mellon University}
\authorblockA{\authorrefmark{2}Field AI\\ Email: striest@andrew.cmu.edu}
}

\maketitle

\begin{strip}
    \centering
    \includegraphics[width=\linewidth]{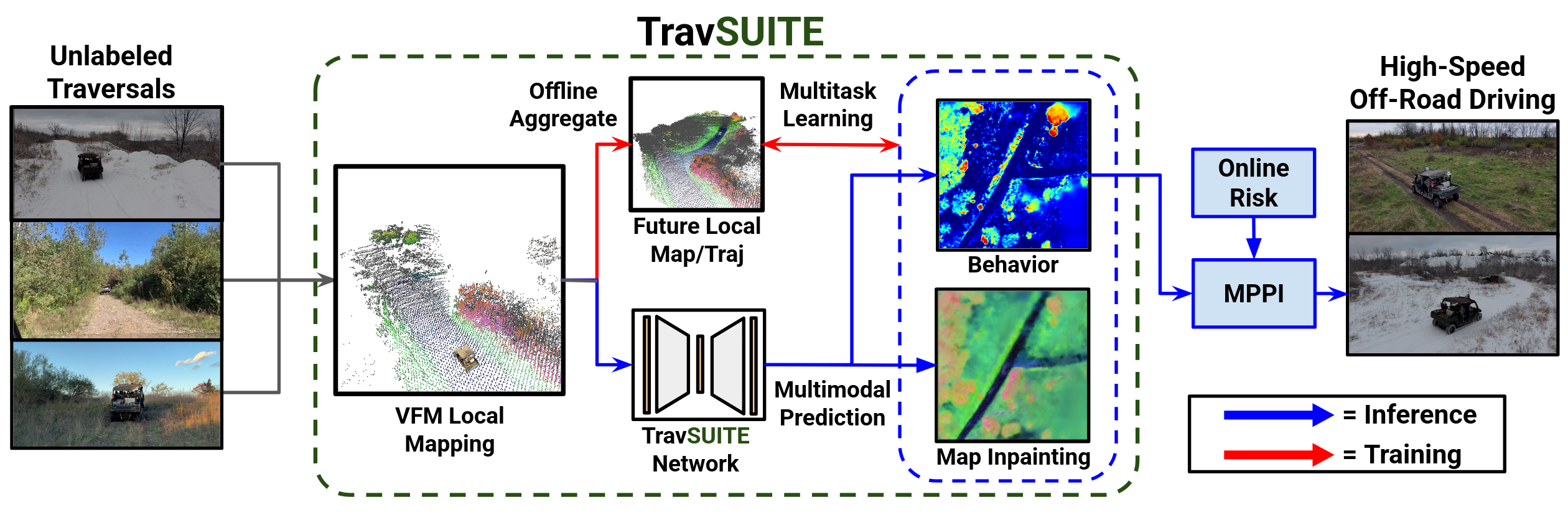}
    \captionof{figure}{TravSUITE is a learning-based off-road traversability system that expands upon prior work in off-road driving by combining off-road perception and planning tasks such as map inpainting \cite{shaban2022semantic, meng2023terrainnet, triest2024unrealnet, zhang2025creste, jung2025uncertainty} and traversability learning \cite{ratliff2009learning, frey2023fast, wigness2018robot, cai2022risk} into a unified, self-supervised framework without any human annotation.}
\end{strip}

\begin{abstract}
Traversability analysis in off-road settings remains a fundamental challenge for mobile robots. Key difficulties include constructing an accurate and expressive local map from multi-modal sensor data, and using the map to design traversability rules that yield desirable navigation behavior. Importantly, this system must be resilient to the limited sensing regime brought about by complex environments and high speeds. In this paper, we present TravSUITE, a traversability system suitable for high-speed navigation in off-road environments. TravSUITE consists of two major components: 1) an efficient voxel mapper that leverages visual foundation models (VFMs) to build a rich geometric-semantic local map from streams of on-board sensor data, and 2) a unified neural network that jointly predicts traversability-relevant quantities in bird's eye view (BEV), including geometry, semantics, speed and cost. Our training strategy is entirely annotation-free and self-supervised, leveraging tasks such as map inpainting and inverse reinforcement learning (IRL) to learn both map representations and traversability. We perform a thorough ablation study and comparison to state-of-the-art approaches, and the results indicate that cost learning and auxiliary inpainting each contribute significantly to planning quality, and their combination is critical for achieving state-of-the-art performance in off-road path planning. We also design a simple risk adaptation mechanism to leverage our method's uncertainty estimates at deploy-time, and demonstrate that a combination of inpainting and risk estimation can result in $80\%$ fewer navigation errors and $5\%$ faster autonomous traversal speeds in real-world hardware experiments.

% I can make the traversal speed increase seem bigger if I include the intervention time but idk if that's right. 

\end{abstract}

\IEEEpeerreviewmaketitle

\section{Introduction}

Many important application domains such as construction \cite{afsari2021fundamentals, bellicoso2018advances}, disaster response \cite{Scherer:2022, tranzatto2022cerberus, agha2021nebula}, forestry \cite{boxan2024fomo, borges2022survey}, and defense \cite{urmson2004high, bagnell2010learning, young2021robot}, require robots to be able to navigate in off-road environments reliably and autonomously. Most practical approaches see a mobile robot 1) use its on-board sensors to build a local map of its surroundings, 2) perform traversability analysis on the local map to determine where it can and cannot traverse, and 3) use the results of its traversability analysis to plan safe paths to a desired goal. This approach is preferred in safety-critical applications due to its modularity, interpretability, and ease of incorporating additional safety constraints, and has thus been the dominant framework for performing navigation across legged \cite{fankhauser2018robust, frey2023fast, tranzatto2022cerberus, bouman2020autonomous} and wheeled platforms \cite{team2005stanford, bagnell2010learning, meng2023terrainnet, maturana2022semantic}, for both learning \cite{castro2022does, frey2024roadrunner, triest2023learning, zhang2025creste, wigness2018robot, tranzatto2022cerberus} and non-learning based \cite{dixit2024step, fankhauser2018robust, chilian2009stereo, krusi2017driving} methods.

While this approach may seem simple at first glance, both the local mapping and traversability problems become very challenging in complex environments such as off-road settings. The quality of the robot's local map can be significantly degraded due to factors such as observation noise and occlusion. Cost function design is a complex function of robot capability, operator preference and the environment's geometry and semantics. Effective robot navigation systems must be able to produce reliable and nuanced traversability estimates even in complex environments where the presence of observational uncertainty may render the local map erroneous or incomplete.

% An ideal traversability analysis system should thus 1) be able to encode enough geometric and semantic nuance from the environment to generate reasonable behavior, 2) be able to detect and compensate for sensing uncertainty, and 3) be efficient at both train and deploy time.

There exist several recent works that aim to improve the resilience of off-road mapping pipelines to observational noise and uncertainty \cite{shaban2022semantic, meng2023terrainnet, triest2024unrealnet, frey2024roadrunner}. These approaches typically rely on using a large neural network to predict some form of ground-truth map representation (derived from privileged data such as hindsight, hand-labeling, simulation, etc.) from the current observations. These approaches have been demonstrated to produce reliable local maps that are robust to occlusions and measurement sparsity at high speeds and long ranges \cite{meng2023terrainnet, frey2024roadrunner}. However, these methods often rely on robot-specific traversability data \cite{frey2024roadrunner}, or environment-specific semantics and hand-annotations \cite{shaban2022semantic, meng2023terrainnet}, making it challenging to generalize these approaches to novel robots or environments.

To address the challenge and tedium of hand-designing traversability rules, there has been significant prior work in learning cost functions directly from robot traversal data \cite{bagnell2010learning, frey2023fast, triest2023learning, castro2022does, karnan2023sterling, triest2024velociraptor, zhang2025creste}. While these approaches have demonstrated promising results across many environments and robot morphologies, there lacks a common consensus on the choice of input representation (e.g. geometric, semantic, visual foundation model (VFM) features, etc.) \cite{castro2022does, karnan2023sterling, frey2023fast, triest2024velociraptor}, and its effect on the downstream traversability are not well-explored. Additionally, these approaches may not account for perceptual uncertainty \cite{ratliff2009learning, castro2022does, karnan2023sterling}, and often lack large-scale datasets and dense supervision signals \cite{frey2023fast, triest2024unrealnet, sivaprakasam2024salon, castro2022does}.

TravSUITE (Fig. 1) synthesizes insights from both families of work to produce reliable and expressive traversability estimates in the presence of observational uncertainty. TravSUITE consists of three major components, 1) a flexible and efficient mapping module that produces real-time geometric-semantic map representations, 2) a framework for generating large, self-supervised, annotation-free perception and behavior datasets from robot experience, and 3) a unified neural network trained to jointly predict inpainted map representations and traversability estimates from this dataset with uncertainty. We leverage the flexibility of our mapping module and learning framework to compare against state-of-the-art approaches \cite{meng2023terrainnet, triest2024velociraptor, zhang2025creste} and systematically evaluate what design decisions (such as mapping representation and training tasks) are important for improving path planning performance. We find that across all of our experiments, the existence of an inpainting task significantly improves traversability estimation performance, regardless of mapping representation. We also demonstrate that learning distributions of cost is important to enabling safer traversal at deploy time.

\section{Related Work}

The vast majority of fieldable off-road systems make use of some local map of the robot's immediate surroundings \cite{team2005stanford, agha2021nebula, meng2023terrainnet, tranzatto2022cerberus, Scherer:2022}. Typically, this local map is typically represented in one of two ways, as bird's eye view (BEV/2.5D), or volumetric (3D). BEV-based approaches elect to partition the robot's surroundings into ``cells'' of finite horizontal ($xy$) extent and either infinite (or hand-designed) vertical extent. Volumetric approaches partition the local environment into cells of finite $xyz$ extent (i.e. voxels) \cite{frey2022locomotion, overbye2022g, alama2025rayfronts}. In either case, cells store navigation-relevant quantities derived from sensor data, such as the elevation of the support surface \cite{fankhauser2016universal, fan2021step}, semantic probabilities \cite{maturana2022semantic, meng2023terrainnet}, appearance data \cite{castro2022does, karnan2023sterling} or deep features \cite{erni2023mem, triest2024velociraptor, jung2025uncertainty}. While BEV-based approaches sacrifice representation fidelity (e.g. overhangs, multiple support surfaces), they are often more computationally efficient and closer represent the manifold that terrestrial robots must reside on. Volumetric approaches, on the other hand, more closely represent the local environment. However, they typically are more challenging to maintain and often require additional support surface estimation techniques for practical use by terrestrial robots \cite{krusi2017driving}.

Similarly, various methods for encoding semantics have been proposed in the literature. The dominant approach for off-road driving is to perform semantic segmentation over a fixed set of traversability-relevant classes (e.g. trails, rocks, etc.) \cite{maturana2018real, jiang2020rellis3d, meng2023terrainnet, goose-dataset}, or traversability classes themselves \cite{guan2022ga, shaban2022semantic}. There has also been recent interest in encoding semantic information via VFMs such as Segment Anything \cite{kirillov2023segment} or Dino \cite{oquab2023dinov2}. These approaches have the potential benefit of additional semantic granularity \cite{triest2024velociraptor}, or open-set segmentation capabilities \cite{weerakoon2025behav}. In either case, both semantic probabilities and VFM features can be aggregated in both BEV and voxel-based approaches \cite{maturana2018real, overbye2022g, erni2023mem, shaban2022semantic, meng2023terrainnet, triest2024velociraptor, zhang2025creste, jung2025uncertainty}. 

There also exists a significant body of work on filling in occlusions in the local map. Early approaches rely on leveraging classical techniques such as Markov random fields \cite{maturana2022semantic, wellington2005interacting} to inpaint the terrain surface. Advances in computer vision have enabled large neural networks to learn BEV-space inpainting on a number of quantities including geometry \cite{triest2024unrealnet, jung2025uncertainty}, semantics \cite{shaban2022semantic, meng2023terrainnet}, costs \cite{frey2024roadrunner, patel2024roadrunner} or VFM features \cite{zhang2025creste}. These approaches typically rely on predicting the map representation at a fixed number of steps in the future. To the authors' knowledge, inpainting is always performed in BEV space and no work currently exists that directly inpaints volumetric representations for off-road environments. 

There also exist many approaches to performing off-road traversability. A popular and straightforward approach is to hand design a cost function that relies on geometric properties of the local environment \cite{fankhauser2018robust, chilian2009stereo, krusi2017driving, fan2021step, triest2024unrealnet} such as the local slope and roughness of the terrain surface. While such an approach has been demonstrated to be effective for certain use cases and austere environments \cite{team2005stanford, miller2006cornell, pfaff2007efficient, agha2021nebula, Scherer:2022}, a well-known failure case of these methods is false positives due to tall but compliant obstacles such as vegetation. Thus, approaches that hand-craft a traversability function based on geometry and semantics have been used to navigate in more challenging environments such as forests and snow \cite{maturana2018real, shaban2022semantic, meng2023terrainnet, frey2024roadrunner}. Another popular family of approaches relies on regressing some sort of traversability signal to the perception representation using deep learning. Many possible quantities have been proposed and demonstrated as a proxy for traversability, including experienced slip or roughness \cite{wellhausen2019should, castro2022does, frey2023fast, cai2023evora, karnan2023sterling}, simulated rollouts \cite{frey2022locomotion}, and probability of expert demonstrations \cite{ratliff2009learning, zhang2018integrating, triest2023learning, zhang2025creste, jung2023v}. Oftentimes, it is valuable to produce a \textit{distribution} of costs to account for uncertainty in the perception representation or traversability function. Uncertainty in the perception can be expressed as a distribution over geometric features \cite{fan2021step} or semantic classes \cite{meng2023terrainnet, shaban2022semantic}. This uncertainty is propagated through, or incorporated in, simpler traversability rules \cite{fan2021step, meng2023terrainnet} to produce a cost distribution. Similarly, uncertainty in the cost function evaluation can also be expressed via reconstruction error \cite{stolzle2022reconstructing, triest2024velociraptor, schmid2022self}, ensembling \cite{ratliff2009learning, triest2023learning}, or evidential learning \cite{cai2023evora}. In all cases, this results in a \textit{distribution} of costs per state that must be compressed into a single value for downstream path planning. It is common practice to use either expected value \cite{ratliff2009learning, meng2023terrainnet}, or conditional value-at-risk (CVaR) \cite{rockafellar2000optimization} to capture long-tail, worst-case costs \cite{fan2021step, cai2023evora, triest2023learning}. 

TravSUITE represents an advance in state-of-the-art for autonomous local navigation in that our framework is flexible with respect to choices of both input representation and semantics. We also design our system to jointly perform map inpainting and traversability learning without any human annotations while maintaining an environment-generic representation. This enables us to perform a systematic evaluation of many design decisions, including input representation, traversability function structure, and use of inpainting tasks. Lastly, our choice of network architecture allows us to produce uncertainty estimates for test-time deployment, and is extensible to other off-road perception tasks, such as roughness/slip prediction \cite{castro2022does, karnan2023sterling, frey2023fast}, which we do not focus on in this work.

\section{Method}

\subsection{Problem Setup}
% consisting of a point cloud $P_t$ and camera image $I_t$ (note that this sensor observation setup can be generalized to many setups including depth images, multi-camera, etc.). 

We will consider a robot tasked with navigating to a distant goal point $x_G$. At every discrete timestep $t$, the robot will have access to a state estimate $x_t$ and observation $O_t$. We will assume we have access to a dataset of expert traversals that are assumed to be locally optimal. A datapoint in this dataset thus consists of three components $[O_{0:t}, x_G, \tau^E]$, which are: the set of observations $O_{0:t}$ that the robot has accrued to this point, a desired goal state $x_G$, and the trajectory that the expert took to reach the goal $\tau^E$. The traversability estimation problem can be thought of as finding a cost function $f_\theta(\tau | O_{0:t}, x_G)$ under which the cost-minimizing trajectory $\hat{\tau}$ closely matches $\tau_E$ under some objective function $\mathcal{L}$ (Eq. \ref{eq:objective}). 

\begin{multline}
    \label{eq:objective}
    \theta^* = \argmin_\theta \mathop{\mathbb{E}}_{O_{0:t}, x_G, \tau^E \sim \mathcal{D}} [\mathcal{L}(\tau^E, \hat{\tau})] \\
    s.t. \quad \hat{\tau} = \argmin_\tau f_\theta(\tau | O_{0:t}, x_G)
\end{multline}

In the scope of this work, we will assume that the cost function $f_\theta(\tau)$ is a weighted combination of distance to the goal $x_G$, and cost incurred from a ``behavior map'' $\Pi_t$. For the purposes of this paper, we will define our behavior map to be in BEV, and contain a cost and maximum speed for every cell (i.e. $\Pi_t = [C_t, S_t], \Pi_t \in \mathbb{R}^{2 \times H \times W}$). By a slight abuse of notation, let $\Pi_t(x)$ represent a function that extracts the relevant values at state $x$ from a map. Given this, we can perform planning by scoring trajectories, conditioned on $\Pi_t$ via Equation \ref{eq:mppi_cost_fn} ($K_1 \approx \infty$ is used to ensure no state in the trajectory exceeds its corresponding speed limit, and $K_2$ is a coefficient used to balance the tradeoff between minimizing costmap cost and final-state distance to goal).

\begin{multline}
    \label{eq:mppi_cost_fn}
    f_\theta(\tau | O_{0:t}, x_G) = \sum_{x \in \tau} \bigg [ C_t(x) + K_1 \mathbbm{1}[||x_v||_2 > S_t(x)] \bigg ] + \\
    K_2 ||x_T - x_G||, \quad K_1 \approx \infty
\end{multline}

\subsection{Decomposing Mapping and Traversability}

Given the form of the cost function in Eq. \ref{eq:mppi_cost_fn}, the goal of the traversability estimation problem is to find the best form and parameters of $f_\theta$ (and thus $\Pi_t$) that yield trajectories $\hat{\tau}$ that minimize the objective $\mathcal{L}(\hat{\tau}, \tau^E)$. It is helpful to decompose the traversability function $f_\theta$ into two components, a mapping function $M_t = h_\psi(O_{0:t})$ that compresses sensor observations into a map $M_t$, and a traversability function $g_\phi(M_t)$ that converts the map into the behavior map $\Pi_t$ (Eq. \ref{eq:trav_and_rep}).

\begin{align}
    \label{eq:trav_and_rep}
    \Pi_t = [C_t, S_t] = (g_\phi \circ h_\psi) (O_{0:t}), \quad \theta = [\phi, \psi] 
\end{align}

It is important to note that the behavior map $\Pi_t$ is produced via a composition of the traversability function $g_\phi$ and mapping function $h_\psi$, which each have their own parameter vector. This is important as many prior works rely on directly improving the mapping function $h_\psi$ through auxiliary tasks such as inpainting \cite{meng2023terrainnet, triest2024unrealnet, zhang2025creste}. In this context, these approaches can be thought of as introducing an auxiliary objective $\mathcal{L}_{aux}$ and dataset $\mathcal{D}_{aux}$ to add additional structure to $M_t = h_\psi(O_{0:t})$. Since the trajectory objective in Eq. \ref{eq:objective} depends on the mapping representation $M$ via Eqs. \ref{eq:trav_and_rep} and \ref{eq:mppi_cost_fn}, it stands to reason that this auxiliary task may be able to improve the downstream traversability estimation. Ultimately, this results in a unified objective (Eq \ref{eq:aux_objective}) in which both the trajectory $\hat{\tau}$ and map $M_t$ are both being optimized.

\begin{multline}
    \label{eq:aux_objective}
    \phi^*, \psi^* = \argmin_{\phi, \psi} \bigg [ \lambda_1 \mathop{\mathbb{E}}_{O_{0:t}, x_G, \tau^E \sim \mathcal{D}} [\mathcal{L}(\tau^E, \hat{\tau}_{\phi, \psi})] + 
    \\ \lambda_2 \mathop{\mathbb{E}}_{O_{0:t}, M^* \sim \mathcal{D}_{aux}} [
        \mathcal{L}_{aux} (M^*, h_\psi(O_{0:t}))
    ] \bigg ]
\end{multline}

\begin{figure}
    \centering
    \includegraphics[width=1.0\linewidth]{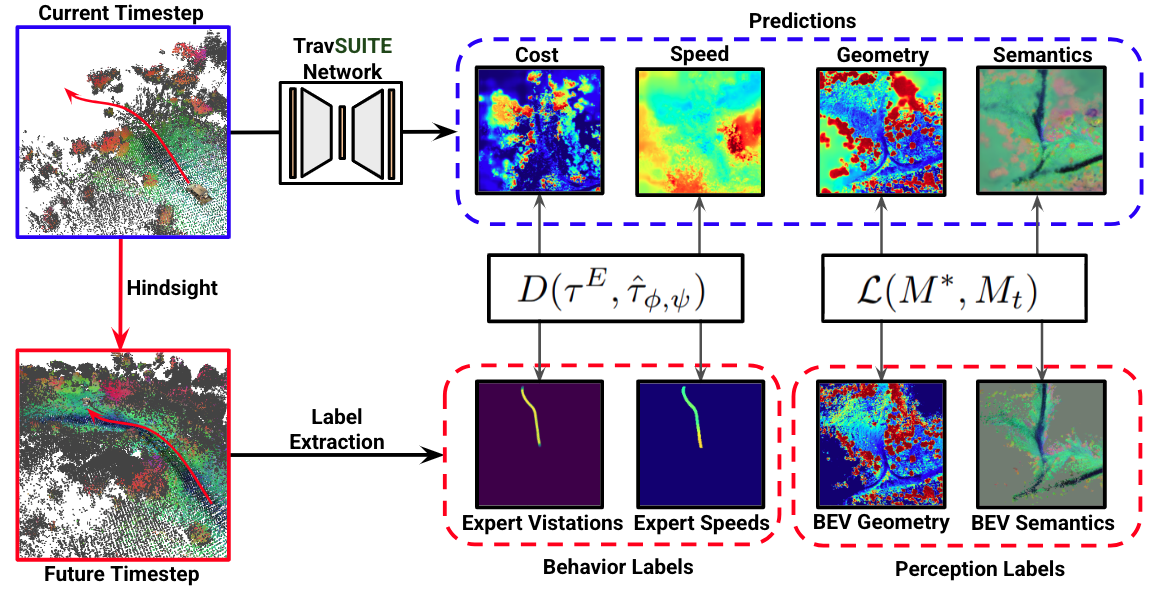}
    \vspace{-3mm}
    \caption{An overview of our supervision. TravSUITE relies entirely on supervision from hindsight, including trajectory and map information. (Blue=derived from current observations, Red=derived from privileged information)}
    \label{fig:supervision_fig}
    \vspace{-5mm}
\end{figure}

Many state-of-the-art traversability approaches can be viewed under this framework \cite{meng2023terrainnet, shaban2022semantic, dixit2024step, zhang2025creste, triest2023learning, karnan2023sterling} through different forms of $g, h$ and whether to supervise $M, \tau$. An overview of our method's supervision can be found in Figure \ref{fig:supervision_fig}. In the following sections, we will describe our choice of supervision for the behavior learning and map learning components of our method.

\subsection{Optimizing Behavior}
In order to perform the behavior learning component of Eq. \ref{eq:aux_objective} (and to generate the behavior labels in Fig \ref{fig:supervision_fig}), we elect to use the MaxEntIRL \cite{ziebart2008maximum} framework, in which the likelihood of a trajectory is inversely proportional to the exponent of its cost, and the objective is to maximize the likelihood of $\tau_E$ (Eq. \ref{eq:maxent_likelihood}). We follow a LEARCH-like procedure \cite{ratliff2009learning}, where we sample an optimal trajectory $\hat{\tau}$ under the current costmap and goal, and then compute the gradient of the IRL objective w.r.t. the costmap to be the difference in state visitations between the expert and learner trajectories ($SV$ is a state visitation extraction function defined in the appendix). This gradient can be used to update $g_\phi$ and $h_\psi$. 

\begin{multline}
    \label{eq:maxent_likelihood}
    \mathcal{L}(\tau^E, \hat{\tau}) \triangleq p(\hat{\tau}) - p(\tau^E), \quad  p(\tau) \propto exp[f_\theta(\tau | O_{0:t}, x_G)] \\ 
    \frac{\partial}{\partial C} D \propto SV(\tau^E) - SV(\hat{\tau})
\end{multline}

We also use the expert trajectory $\tau_E$ to learn a speedmap $S_t$, which contains the maximum safe traversal speed for a given cell. It is important to note that we must produce an upper bound, making it necessary to predict distributions of speed rather than just the mean value. As such, we follow the example used by Cai et al. \cite{cai2023evora} by parameterizing our speedmap $S_t$ as a categorical distribution over fixed bins, per cell. We train our model to minimize squared earth mover's distance to a delta distribution centered on the expert speed for every cell that the expert traversed. In order to encourage safe behavior, we also add a small amount of negative samples (where the target is 0) for untraversed cells (Eq. \ref{eq:speed_loss}). 

\begin{multline}
    \label{eq:speed_loss}
    \mathcal{L}_S = \sum_{x \in \tau^E} [EMD^2(I(x, S_t)), \delta(||x_v||_2))] + \\
    \lambda_s \sum_{x \notin \tau^E} [EMD^2(I(x, S_t)), \delta(0))]
\end{multline}

In practice, we obtain labels for this behavior learning by extracting fixed-length trajectory segments from our dataset of expert traversals. For each datapoint, we sample a trajectory segment $50m$ in length and assign $x_G$ to be the final state in the trajectory. This approach is preferred to time-based sampling as it ensures that demonstrations in more challenging scenarios (with slower expert speeds) do not have reduced path complexity. In order to sample learner trajectories $\hat{\tau}$, we use MPPI \cite{williams2017information} under a kinematic bicycle model and the cost function from Eq. \ref{eq:mppi_cost_fn}. Note that for training stability, we set the speed penalty ($K_1$) to 0 at train time.

\subsection{Optimizing the Perception Representation}
In order to instantiate the perception learning component of Eq. \ref{eq:aux_objective} (and the perception labels in Fig \ref{fig:supervision_fig}), we must define $M^*_t$. Similar to prior work \cite{meng2023terrainnet, zhang2025creste}, we use hindsight supervision in order to generate labels $M^*_t$, and train our mapping network $M_t = h_\psi(O_{0:t})$ to predict $M^*_t$ (Eq. \ref{ref:eq_inpaint}). Note that this objective encourages inpainting as  $M^*_t$ is a function of both past and future timesteps (e.g. $M^*_t \sim O_{0:T}$).

\begin{equation}
    \label{ref:eq_inpaint}
    \mathcal{L}_{aux} = ||M^* - M_t||, \quad M^* \sim O_{0:T}, \quad M_t \triangleq h_\psi(O_{0:t})
\end{equation}

\section{Implementation Details}

% In order to instantiate the algorithm outlined in the previous section, we need to make a number of design decisions with respect to the structure of our mapping and traversability functions, and choice of supervision for both expert demonstrations and inpainting. We outline our major design decisions the the following section and Figure \ref{fig:supervision_fig}. 

\subsection{Aggregating Observations with VFM Voxel Mapping}

\label{sec:voxel_mapping}

In practice, it is not feasible to maintain and process an ever-growing list of observations $O_{0:t}$ when $t$ can span thousands of timesteps or more. To address this, we developed a voxel mapper that aggregates visual and geometric features in a fixed volume around the robot, given state estimates $x_t$. We chose to use a voxel grid as our core mapping representation instead of a BEV-based approach in order to enable our system to be more generically used across environments and tasks (e.g. indoor environments, manipulation), which may require 3D information. Furthermore, we are able to generate an accurate BEV representation from a voxel grid, while the opposite is not true. We also find that our voxel mapper's processing times are competitive with an elevation mapping-based baseline and is sufficient for real-time operation. Further details are provided in the appendix. For every timestep in our expert dataset, we are able to represent $O_{0:t}$ with either a voxel map $V_t$ or BEV map $B_t$ containing much of the traversability-relevant information from all prior observations. These maps $V_t, B_t$ will serve as input to the learned components of TravSUITE. For all experiments, our map size is $100m\times100m\times20m$ at $0.4m\times0.4m\times0.1m$ per cell, centered on the robot. The list of map features can be found in Table \ref{tab:bev_features}.

\begin{table}[h]
    \centering
    \begin{tabular}{c||c|c|c}
        Feature & Modality & Rep. & Description \\
        \hline
        Terrain & Geom. & BEV & Elevation of the terrain surface \\
        Slope & Geom. & BEV & Slope of the terrain surface \\
        Diff & Geom. & BEV & Max elevation over the terrain \\
        min/max $z$ & Geom. & Voxel & The min/max height in each voxel \\
        Seg Logit x16 & Seg. & Both & Class score from Talk2Dino \cite{barsellotti2025talking} \\
        VFM x32 & VFM & Both & VFM feature from Radiov3 \cite{ranzinger2023radio} \\
    \end{tabular}
    \caption{List of Mapping features}
    \label{tab:bev_features}
    \vspace{-5mm}
\end{table}

% Many state-of-the-art traversability approaches can be viewed under this framework (Table \ref{tab:method_positioning}) through different choices of $g, h$ and whether to supervise $M, \tau$.

% \begin{table}[]
%     \centering
%     \begin{tabular}{c||c|c|c|c}
%         Method & learn $g$? & learn $h$? & $\mathcal{L}(\tau)$? & $\mathcal{L}(M)$? \\
%         \hline
%         STEP \cite{fan2021step} & no & no & no & no\\
%         TerrainNet \cite{meng2023terrainnet} & no & yes & no & yes \\
%         Velociraptor \cite{triest2024velociraptor} & yes & no & yes & no \\
%         Creste \cite{zhang2025creste} & yes & yes & yes \footnote[1]{not cotrained} & yes \footnote[1]{not cotrained} \\
%     \end{tabular}
%     \caption{Comparison of prior work}
%     \label{tab:method_positioning}
% \end{table}

\subsubsection{Generating Hindsight Supervision from Voxel Maps}

In order to perform inpainting, we construct an oracular voxel map $(M^*_t)$ and BEV map ($B^*_t$) for every timestep in our dataset. We do this in a similar fashion to TerrainNet \cite{meng2023terrainnet} by building a voxel map $V^*_t$ from both past and future sensor measurements. Instead of setting a fixed time window to aggregate (e.g. $V^*_T \sim [V_{t-k} \dots V_{t+k}]$), we instead keep track of the mapping volumes for each timestep and take as $V^*_t$ the aggregate of all voxels from every voxel grid that intersects with the volume defined by $V_t$ (detailed algorithm in appendix). This has an advantage in that the quality of aggregation does not depend on robot speed. Additionally, since we directly predict the future map (instead of semantic annotations), this process is fully annotation-free. For both voxel and BEV-based approaches, we choose to only use $B^*$ as a prediction target. Since the supervision is derived from local perception, cells in $B^*$ may still be unreliable at longer sensing distances. To address this, we mask out BEV cells that were observed at a distance no closer than $10m$ from the robot. Note that this distance includes all future poses $x_{0:T}$, meaning $B^*_t$ can contain labels at any distance from the current state $x_t$ and thus requires the network to inpaint.

\subsection{Network Architecture}

\begin{figure}
    \centering
    \includegraphics[width=1.0\linewidth]{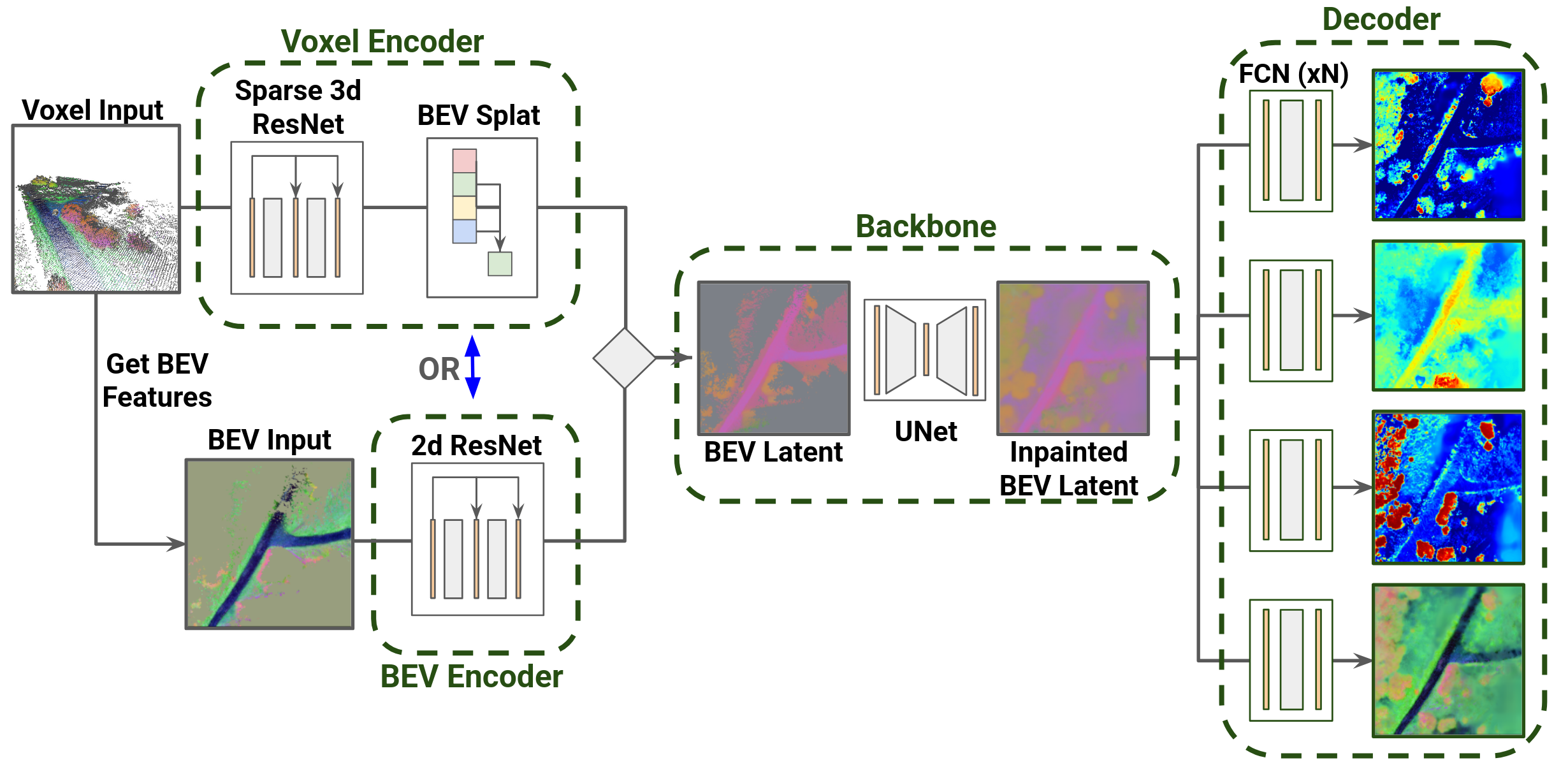}
    \vspace{-3mm}
    \caption{An overview of our network architecture. The TravSUITE network first starts with a ResNet-style encoder (either 2d or 3d depending on the input modality), followed by BEV-space inpainting and parallel BEV decoder FCN heads. Since the individual prediction heads are relatively lightweight, the backbone and encoder must handle the majority of the feature extraction.}
    \label{fig:net_arch}
    \vspace{-5mm}
\end{figure}

We instantiate the perception and traversability function from Equation \ref{eq:trav_and_rep} as a multi-headed neural network, similar to ParaDrive \cite{weng2024drive}. The network takes as input $M_t$ (which may be either $V_t$ or $B_t$), and uses an encoder and backbone to produce a latent BEV representation $Z_t$. Multiple different heads are then used to produce navigation-relevant quantities (e.g. costmaps, speedmaps, semantic maps) from $Z_t$. A high-level architecture diagram is presented in Figure \ref{fig:net_arch}.

\subsubsection{Encoder and Backbone}

The role of the encoder and backbone is analogous to the mapping function $h_\psi$ in the previous section, where the goal is to convert the input into a BEV-space feature tensor of a given channel dimension (e.g. $h_\psi(O_{0:t}) \in \mathbb{R}^{F \times W \times H}$). If the input is $B_t$, this process is simple and can be implemented as a simple ResNet-style \cite{he2016identity} network. If the input is $V_t$, we use a 3D sparse convolution-based analogue using spconv \cite{spconv2022} and then splat the 3D features into BEV using a channel-wise max pool.

The backbone of our network is a UNet \cite{ronneberger2015u}. This UNet comprises the bulk of the learnable parameters in our network and enables the network to attend to longer-range dependencies in the data. The output of the UNet is a feature tensor $Z \in \mathbb{R}^{F \times W \times H}$ and is used as a shared input across all prediction heads $g^i_\phi$. 

\subsubsection{Prediction Heads}

In order to convert $Z$ into the same semantic space as $B^*$, another lightweight decoder FCN \cite{long2015fully} $g_\phi(Z_t)$ is used. Following TerrainNet's example, we use a separate head for each feature modality (e.g. geometry, semantics). A similar decoder architecture is used to predict cost and speed. However, as we are interested in estimating distributions of cost and speed, we parameterize the outputs of these heads as a categorical distribution over fixed bins. This approach has been demonstrated to be useful when the support of the target variable is known (e.g. traction, speed) \cite{cai2023evora, triest2024velociraptor}. This also enables risk-aware behavior via sampling different risk measures of the resulting distributions (Fig. \ref{fig:cost_head_arch}).

\subsection{Leveraging Uncertainty}
\label{sec:online_cvar}

\begin{figure}
    \centering
    \includegraphics[width=0.7\linewidth]{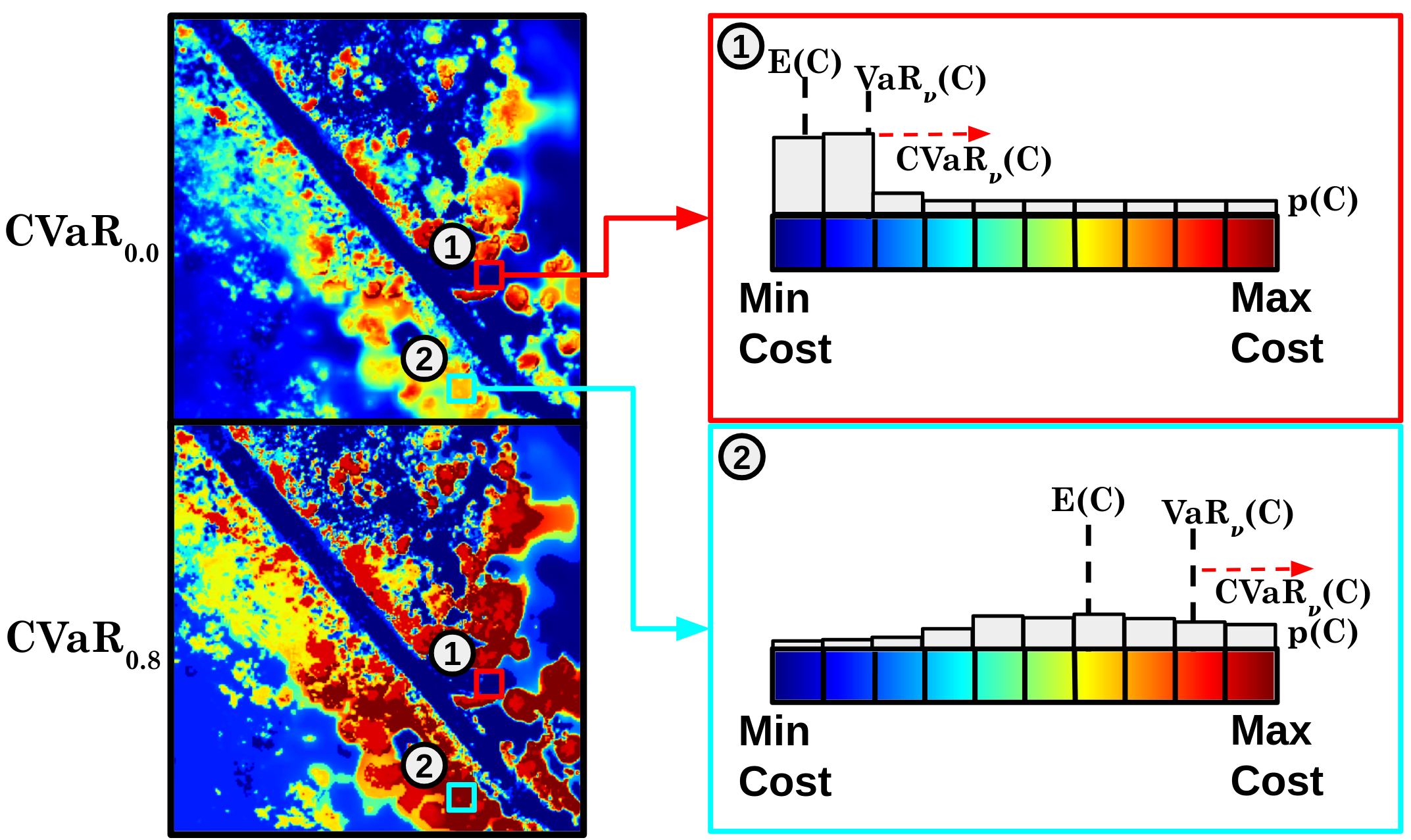}
    \caption{A diagram of the cost architecture and uncertainty mechanism. (Top-Left) A sample costmap at CVaR=0. (Bottom-Left) The same costmap at CVaR=0.8. (Right) A zoom-in of the cost distribution for two select cells. The red cell has a low-entropy cost distribution and thus stays at a similar value for all CVaR values. The cyan cell has a high-entropy cost distribution and has its cost inflated significantly under higher CVaR.}
    \label{fig:cost_head_arch}
    \vspace{-5mm}
\end{figure}

Since both predicted cost and speed are parameterized as categorical distributions, we are able to evaluate various risk measures such as VaR (quantiles) and CVaR (Fig \ref{fig:cost_head_arch}). At test time, we are able to compute a risk-aware cost by evaluating the CVaR of every cell in the costmap. Similarly, we are able to tune the aggressiveness of our speedmap by choosing different quantiles of the resulting speed distributions. 

We also implemented a very simple test-time CVaR modulation technique for costs in which a running risk value $\nu_t$ was maintained and updated based on the current speed. If the current vehicle speed exceeded a set value, or fell below another set value, the risk tolerance was slightly decremented or incremented, respectively (Equation \ref{eq:cvar_tune}). While simple, this method has the desirable behavior of encouraging safe behavior if the robot is moving quickly, but allowing riskier behavior if little navigation progress is being made.

\begin{flalign}
    \label{eq:cvar_tune}
    \nu_t = \begin{cases}
        \nu_{t-1} - \epsilon & v_t < v_{low} \\
        \nu_{t-1} + \epsilon & v_t > v_{high} \\
        \nu_t & \hbox{otherwise}
    \end{cases}
\end{flalign}

\subsection{Training Details}
For all experiments, we leverage a dataset of roughly four hours of expert driving at a primary and secondary test site, across all four seasons using an full-scale, autonomous off-road vehicle equipped with the sensor suite from TartanDrive 2.0 \cite{sivaprakasam2024tartandrive} (of which we use the LiDAR, IMU and RGB camera). In order to generate state estimates $x_t$, we leverage an off-the-shelf Lidar-inertial odometry algorithm \cite{zhao2021super}. Given these state estimates, we are able to run our VFM voxel mapper to generate inertial-frame VFM voxel maps at $10hz$. The dataset contains roughly 100,000 individual frames with a timestep of $0.1s$, split roughly 80/20 between the two locations. The dataset was divided into a train set and two separate test sets. The \textbf{Train} and \textbf{Test 1} sets were partitioned (roughly 60/40 by GPS region) from the primary test site, while \textbf{Test 2} contains the entirety of the data from the secondary site. Experiments were run for five epochs with a batch size of 3. Since the BEV inpainting targets are multi-modal (e.g. geometric/VFM/semantic features), we follow the procedure used in TerrainNet \cite{meng2023terrainnet}, where each modality is weighted separately. Smooth-L1 \cite{smoothl1} ($\beta=0.2$) is used for continuous features, while cross-entropy is used for segmentation features (Eq. \ref{eq:inpainting_loss}).

\begin{multline}
    \label{eq:inpainting_loss}
    \mathcal{L}_M(B^*, B) = \lambda_{geom} L1(B^*_{geom}, B_{geom}) + \\ L1(\lambda_{VFM} ||B^*_{VFM} - B_{VFM}||) + \lambda_{seg} CE(B^*_{seg}, B_{seg})
\end{multline}
    
\section{Experiments}

In order to evaluate what design decisions matter for downstream path planning, we performed a series of rigorous online and offline evaluations using our dataset and real-world navigation trials.

\subsection{Offline Experiments}
In our offline experiments, we answer these questions:

\begin{enumerate}
    \item \textbf{RQ1:} Do auxiliary tasks such as inpainting improve downstream path planning and if so which?
    \item \textbf{RQ2:} How much does the choice of input representation affect downstream performance?
    \item \textbf{RQ3:} How do we compare to relevant baselines?
\end{enumerate}

Across our offline experiments, we use a common set of behavior and inpainting metrics. In order to measure the efficacy of the behavioral learning, we report two metrics. Modified Hausdorff distance \cite{dubuisson1994modified} (MHD) between the expert trajectory and optimal learner trajectory is used to evaluate the cost prediction component. The average probability of the expert speed under the learned speedmap is used to evaluate the speedmap prediction component. To measure inpainting performance, we can directly compute the error between $h_\psi(O_{0:t})$ and $B^*_t$. We compute error separately for each feature modality as:

\begin{enumerate}
    \item \textbf{Geometry}: $L_1$ error.
    \item \textbf{Semantics}: $L_1$ distance between the semantic distributions (note that neither distribution is one-hot and this essentially measures total variation distance).
    \item \textbf{VFM}: $L_2$ error.
\end{enumerate}

% In order to have better insight as to the effects of our design decisions, we hand-annotated data points from the two test sets via the following criteria:
% \begin{enumerate}
%     \item \textbf{Trail}: whether the expert demonstration was on a visually discernible trail. This was either true or false
%     \item \textbf{Obstacle density}: the amount of obstacles in the vicinity of the expert demonstration. This was partitioned into three categories. \textbf{Open}: 0-3 obstacles within 5m of the expert trajectory, \textbf{Sparse}: 3+ obstacles within 5m of the expert trajectory, or \textbf{Dense}: 3+ obstacles where the expert trajectory required contact with the surrounding environment.
% \end{enumerate}
% This resulted in six categories of demonstrations roughly organized by navigational difficulty, enabling more fine-grained analysis of our approach.

\subsection{RQ1: The Effect of Inpainting on Behavior Learning}

% Compare velociraptor to VFM voxel inpaint. Also TODO more granularity on inpainting types

In order to answer \textbf{RQ1}, we trained several variations of our voxel-based network using different permutations of inpainting modalities. In particular, we compare using no inpainting task (\textbf{None} in Table \ref{tab:inpaint_ablation}), inpainting only geometric/VFM/segmentation features, and cotraining all inpainting modalities together. Overall we find that across both datasets, inpainting improves both the costmap and speedmap components of our network. Notably, all ablations in this experiment had the same number of behavioral examples, confirming the hypothesis that learning a better map represenation $M$ results in better planning behavior. Perhaps unsurprisingly, using more modalities together seems to result in the best performance, as traversability is a function of both geometry and semantics. 

\begin{table}[]
    \centering
    \begin{tabular}{c|c||c|c}
        Inpainting & Dataset & MHD $\downarrow$ & Speed Prob. $\uparrow$ \\
        \hline
        All & Test 1 & \textbf{1.3936} & 0.1159 \\ 
        Geometry & Test 1 & 1.4138 & 0.1054 \\ 
        VFM & Test 1 & 1.5400 & \textbf{0.1162} \\ 
        Segmentation & Test 1 & 1.4877 & 0.1120 \\ 
        None & Test 1 & 1.7744 & 0.1026 \\ 
        % Velociraptor \cite{triest2024velociraptor} & Test 1 & 1.5246 & 0.0967 \\
        \hline 
        All & Test 2 & \textbf{1.3988} & \textbf{0.1258} \\ 
        Geometry & Test 2 & 1.4410 & 0.0982 \\ 
        VFM & Test 2 & 1.4619 & 0.1158 \\ 
        Segmentation & Test 2 & 1.4228 & 0.1004 \\ 
        None & Test 2 & 1.5400 & 0.1086 \\
        % Velociraptor \cite{triest2024velociraptor} & Test 2 & 1.4961 & 0.1017 \\
    \end{tabular}
    \caption{Effect of inpainting on planning tasks}
    \label{tab:inpaint_ablation}
    \vspace{-3mm}
\end{table}

\subsection{RQ2: Choice of Input Representation}

In order to answer \textbf{RQ2}, we explore several choices of input representations to our method. In particular, we consider three types of inputs: 

\begin{enumerate}
    \item \textbf{Voxel}: $V_t$ from Section \ref{sec:voxel_mapping}.
    \item \textbf{BEV}: $B_t$ from Section \ref{sec:voxel_mapping}.
    \item \textbf{Voxel 1-frame}: $V_t$, but only containing voxels from the current timestep (to ablate the importance of aggregating observations, as some approaches that inpaint do not aggregate measurements \cite{philion2020lift, zhang2025creste}, or find performance gains by not doing so \cite{shaban2022semantic, meng2023terrainnet}).
\end{enumerate}

Similar to the previous section, we report the planning metrics on the held-out dataset in Table \ref{tab:input_rep_ablation}. Additionally, we report the inpainting accuracy in Table \ref{tab:input_rep_ablation_inpaint}. Note that the cells used for With respect to path-planning metrics, we observe that the voxel method performs better on the costmap metric while the BEV method is better with respect to the speedmap metric. This indicates that (at least for our environments), both BEV and 3D representations perform reasonably well for traversability. Across both datasets, the single-frame voxel input method performs the worst, suggesting that significant performance gains can be achieved by aggregating sensor measurements.

We find again that aggregation of observations results in better inpainting performance for both observed and unobserved cells. This corroborates the findings of Shaban et al. \cite{shaban2022semantic} in that temporal aggregation is helpful in simplifying the map prediction task when odometry is reliable. We observe that with aggregation, both BEV and voxel-based inputs perform similarly, with the BEV-based model performing marginally better. This is likely due to the labels being generated using the same process as the inputs, making the BEV to BEV learning task somewhat simpler.

\begin{table}[]
    \centering
    \begin{tabular}{c|c||c|c}
        Input & Dataset & MHD $\downarrow$ & Speed Prob. $\uparrow$ \\
        \hline
        Voxel & Test 1 & \textbf{1.3936} & 0.1159 \\
        BEV & Test 1 & 1.4440 & \textbf{0.1184} \\
        Voxel no agg. & Test 1 & 1.5061 & 0.1109 \\
        \hline
        Voxel & Test 2 & \textbf{1.3988} & 0.1258 \\
        BEV & Test 2 & 1.4373 & \textbf{0.1311} \\
        Voxel no agg. & Test 2 & 1.4452 & 0.1031 \\
    \end{tabular}
    \caption{Effect of input representation on planning tasks}
    \label{tab:input_rep_ablation}
    \vspace{-3mm}
\end{table}

% \begin{table*}[]
%     \centering
%     \begin{tabular}{c|c||c|c|c|c|c|c|c|c|c}
%         Input & Dataset & Geom. Inp. & Geom Outp. & Geom Total & VFM Inp. & VFM Outp. & VFM Total & Seg Inp. & Seg Outp. & Seg Total \\
%         \hline
%         Voxel & Test 1 & 0.4277 & \textbf{1.3306} & 0.7253 & 1.0302 & 1.8356 & 1.2324 & 0.2536 & \textbf{0.4523} & 0.3087 \\
%         BEV & Test 1 & 0.4016 & 1.3492 & 0.7137 & 0.9182 & 1.8285 & 1.1450 & 0.2441 & 0.4558 & 0.3019 \\
%         Voxel No Agg. & Test 1 & 1.5143 & 1.7268 & 1.5952 & 1.9472 & 1.9644 & 1.9533 & 0.4937 & 0.5281 & 0.5064 \\
%         \hline
%         Voxel & Test 2 & 0.3863 & 1.0111 & 0.5221 & 1.0869 & 1.9191 & 1.2389 & 0.3891 & 0.5703 & 0.4279 \\
%         BEV & Test 2 & \textbf{0.3502} & \textbf{0.9077} & \textbf{0.4704} & \textbf{0.9667} & \textbf{1.8781} & \textbf{1.1313} & \textbf{0.3645} & \textbf{0.5353} & \textbf{0.4019} \\
%         Voxel No Agg. & Test 2 & 1.6422 & 1.4921 & 1.6326 & 2.0510 & 2.0147 & 2.0331 & 0.5878 & 0.6244 & 0.6033\\
%     \end{tabular}
%     \caption{Effect of input representation on inpainting tasks TODO get more condensed numbers}
%     \label{tab:input_rep_ablation_inpaint}
% \end{table*}

\begin{table}[]
    \centering
    \begin{tabular}{c|c||c|c|c}
        Input & Dataset & Geom $\downarrow$ & VFM $\downarrow$ & Seg $\downarrow$ \\
        \hline
        Voxel & Test 1 & 0.7253 & 1.2324 & 0.3087 \\
        BEV & Test 1 & \textbf{0.7137} & \textbf{1.1450} & \textbf{0.3019} \\
        Voxel No Agg. & Test 1 & 1.5952 & 1.9533 & 0.5064 \\
        \hline
        Voxel & Test 2 & 0.5221 & 1.2389 & 0.4279 \\
        BEV & Test 2 & \textbf{0.4704} & \textbf{1.1313} & \textbf{0.4019} \\
        Voxel No Agg. & Test 2 & 1.6326 & 2.0331 & 0.6033\\
    \end{tabular}
    \caption{Effect of input representation on inpainting tasks}
    \label{tab:input_rep_ablation_inpaint}
    \vspace{-3mm}
\end{table}

\subsection{RQ3: Comparison to Other Methods}
\label{sec:baselines}

We also compare our method to several approaches in recent literature. In order to do so, many of the ablations made use of an inpainting-only backbone. This backbone shared the architecture described in the previous section, but was trained only on the inpainting task. \textbf{TerrainNet Cfn.}: In order to produce a TerrainNet-style \cite{meng2023terrainnet} baseline, we used the predicted geometric and semantic maps to implement the hand-designed cost function prescribed by TerrainNet. In order to improve the performance, we fine-tuned the coefficients of the cost function with IRL. As TerrainNet does not produce a speedmap, we do not report a speed metric. \textbf{Cascaded Trav.}: We produced a Creste-style \cite{zhang2025creste} baseline by training a FCN with IRL, using the predicted geometric and VFM maps as input. \textbf{FCN no Inpaint}: We also trained a Velociraptor-style \cite{triest2024velociraptor} baseline by training a lightweight FCN on $B_t$, using only the behavioral objective. \textbf{Parallel Trav.} is equivalent to inpainting with all modalities from Table \ref{tab:inpaint_ablation}. Results can be found in Table \ref{tab:baselines}.

Overall, we find that both the parallel and cascaded traversability methods outperform the TerrainNet baseline, suggesting that a more complex cost function can lead to improved navigation performance. Interestingly, we find that the FCN baseline (the only method that does not use inpainting) performs the worst across both metrics, suggesting that inpainting with larger architectures is important to learning effective navigation maps. Additionally, we note that the FCN baseline outperforms the larger architecture (no inpainting from Table \ref{tab:inpaint_ablation}), suggesting that it is difficult to train a large-scale architecture with traversability supervision alone. The performance of the parallel and cascaded architecture is relatively comparable with the former producing better speedmaps, and the latter producing better costmaps. Overall, this suggests that learning traversability from demonstrations can outperform traditional baselines, but it is important to add an inpainting task. Some qualitative prediction examples are presented in the hardware experiments section.

\begin{figure*}
    \centering
    \includegraphics[width=0.8\linewidth]{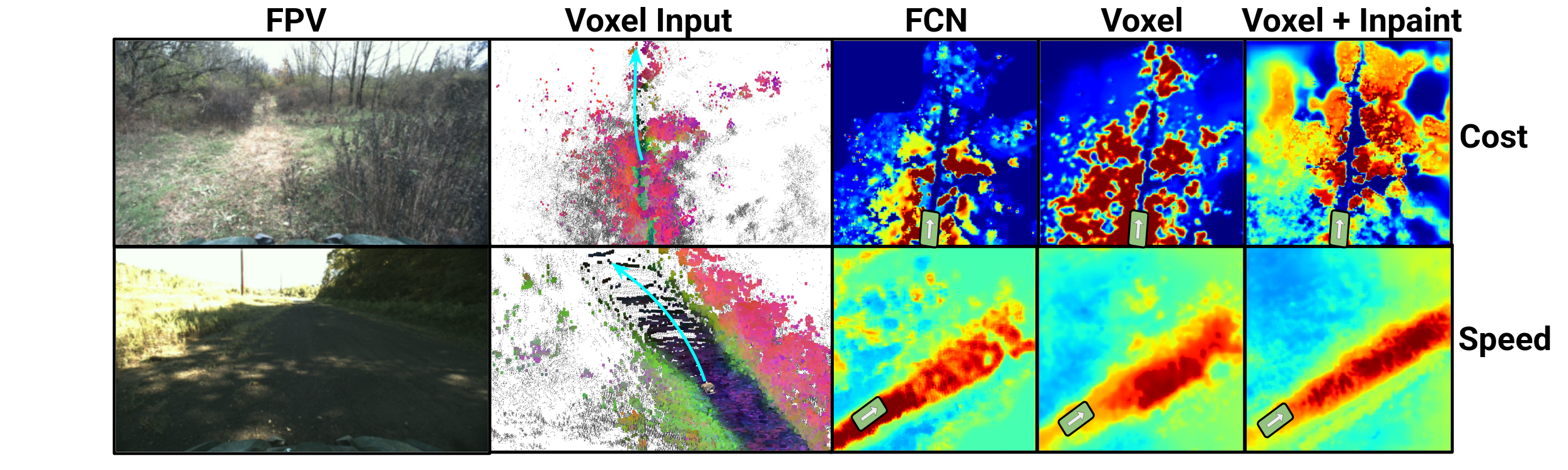}
    \vspace{-3mm}
    \caption{Two scenarios from the test set. Cost and speed maps are cropped and enlarged to highlight regions of interest. (Top row): The robot must navigate through a path obscured by dense vegetation (cyan arrow). The Voxel+Inpaint network does the best at assigning low cost to the trail and high cost to the surrounding unobserved space, which is likely vegetation. (Bottom row): The robot is traveling at high speed on an open trail. Due to the speed, LiDAR measurements are relatively sparse. The inpainting method is able to fill in the missing trail features and keep a consistent speed map.}
    \label{fig:offline_qual}
    \vspace{-5mm}
\end{figure*}

Sample qualitative results are presented in Figure \ref{fig:offline_qual} in which we compare the voxel network with inpainting (Voxel+All in Table \ref{tab:inpaint_ablation}) to two networks without inpainting (Voxel+None in Table \ref{tab:inpaint_ablation} and FCN no Inpaint from Table \ref{tab:baselines}). We observe significant improvements in the stability and accuracy of both the cost and speed maps as a result of inpainting. In practice, this prediction stability is important for real-world navigation.

\begin{table}[]
    \centering
    \begin{tabular}{c|c||c|c}
        Approach & Dataset & MHD $\downarrow$ & Speed Prob. $\uparrow$ \\
        \hline
        Parallel Trav. & Test 1 & 1.3936 & \textbf{0.1159} \\
        Cascaded Trav. & Test 1 & \textbf{1.3043} & 0.1076 \\
        TerrainNet Cfn. & Test 1 & 1.4553 & N/A \\
        FCN no Inpaint & Test 1 & 1.5246 & 0.0967 \\
        \hline 
        Parallel Trav. & Test 2 & 1.3988 & \textbf{0.1258} \\
        Cascaded Trav. & Test 2 & \textbf{1.3845} & 0.1156 \\
        TerrainNet Cfn. & Test 2 & 1.3965 & N/A \\
        FCN no Inpaint & Test 2 & 1.4961 & 0.1017 \\
    \end{tabular}
    \caption{Comparison to similar methods}
    \label{tab:baselines}
    \vspace{-3mm}
\end{table}

\subsection{Hardware Experiments}

We also perform several real-world hardware experiments to answer these questions:

\begin{enumerate}
    \item Does our approach enable off-road navigation?
    \item Does risk-aware planning enable safer navigation?
    \item Do our offline findings translate to hardware?
\end{enumerate}

In order to evaluate our algorithm on physical hardware, we set up a navigation trial using our full-scale ATV at our primary test site. The ATV was tasked with navigating as quickly as possible to a series of waypoints designed to force the vehicle to interact with a mixture of on-trail and off-trail scenarios, as well as vegetation and man-made obstacles. The experiment took place in the winter with somewhat significant snow cover. A full traversal of the course was roughly 1.4km.

% \begin{figure}
%     \centering
%     \includegraphics[width=\linewidth]{figs/nav_trial_collage.png}
%     \caption{Example images from the navigation trial. The robot was required to contend with a number of challenging scenarios including tall vegetation, man-made obstacles, snow, adverse sensing, etc.}
%     \label{fig:nav_trial_collage}
% \end{figure}

A more complex version of MPPI was used on hardware. This version of MPPI used a kinematic bicycle model with additional throttle and steering dynamics to model the transient behavior of the throttle and steering actuators on the platform. In addition to the cost function from Equation \ref{eq:mppi_cost_fn}, an additional lateral acceleration limit was added. Given the challenging conditions, the mean predicted speed from the speedmap prediction was used. All mapping and traversability computation was run at roughly $5hz$ using an on-board laptop with an 13th gen Intel i9 CPU and Nvidia 4090 laptop GPU.

Several ablations from the previous section were compared:

\begin{enumerate}
    \item \textbf{Parallel Trav.}: from Section \ref{sec:baselines}. 
    \item \textbf{TerrainNet Cfn.}: from Section \ref{sec:baselines}. Note that we also trained a speed prediction head from the inpainted latent features in a similar fashion to the previous sections.
    \item \textbf{FCN no Inpaint}: from Section \ref{sec:baselines}.
    \item \textbf{Parallel Trav. + CVaR}: \textbf{Parallel Trav.}, but also using the online CVaR selection method from the Section \ref{sec:online_cvar}
\end{enumerate}

We use the following metrics for evaluation:

\begin{enumerate}
    \item \textbf{Number of Interventions (Int.)}: Number of times a safety operator was required to correct the robot. The safety operator was instructed to intervene if and only if 1) a collision with an obstacle appeared imminent, or 2) the robot stopped making progress towards the next waypoint. If an intervention was called, the safety operator was instructed to teleoperate the vehicle past the offending obstacle and then resume autonomy.
    \item \textbf{Number of Collisions (Col.)}: Of the interventions above, the amount that were due to an obstacle collision. We report this number separately as many interventions due to lack of progress can be addressed with better backtracking and planning, while collisions cannot.
    \item \textbf{Average/Top Auto. Speed}: The average/top speed while autonomous.
\end{enumerate}

% I measured the course at ~4800ft = 1.46km

\subsection{Analysis of Hardware Results}

\begin{figure}
    \centering
    \includegraphics[width=0.85\linewidth]{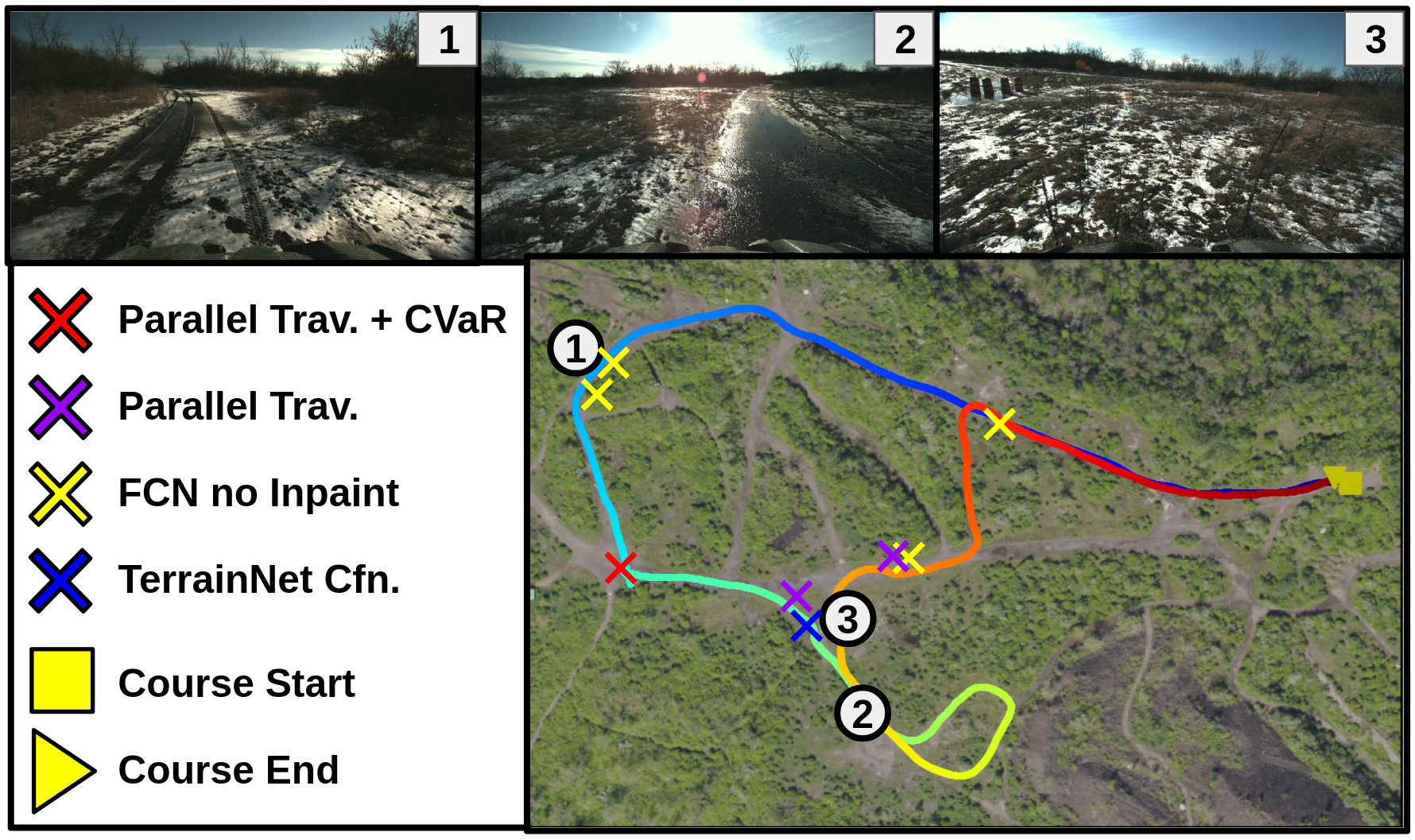}
    \caption{GPS map of our navigation course. The course totaled about 1.4km in length and contained a number of challenges including snow, vegetation and man-made obstacles. In order to successfully navigate the course, the vehicle must be able to avoid OOD obstacles, drive through tall vegetation, and intelligently select traversal speed based on the terrain surface.}
    \label{fig:nav_cousrse_gps}
\end{figure}

\begin{figure}[]
    \centering
    \includegraphics[width=0.85\linewidth]{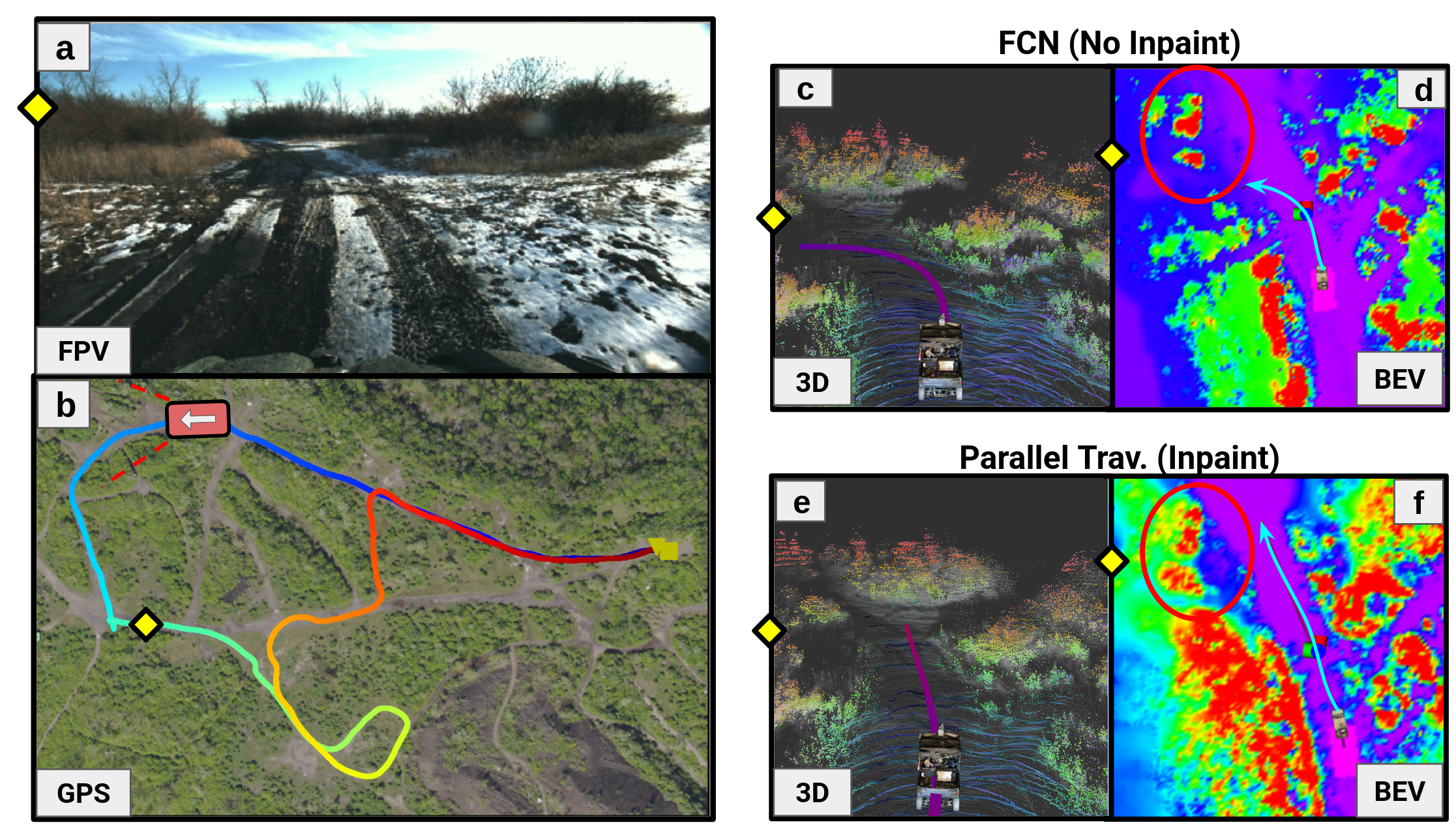}
    \caption{Qualitative figure from the navigation trial. (a, b) After reaching the current waypoint, the ATV must navigate to another waypoint about $150m$ to its left. (c, d) Without inpainting, the FCN baseline underestimates the cost of the vegetation out of the camera view, resulting in an intervention. (e, f) The inpainting baseline is able to successfully estimate the cost of the vegetation, thus encouraging the local planner to take a longer path on the trail.}
    \label{fig:nav_trial_1_qual}
\end{figure}

\begin{figure}[]
    \centering
    \includegraphics[width=0.95\linewidth]{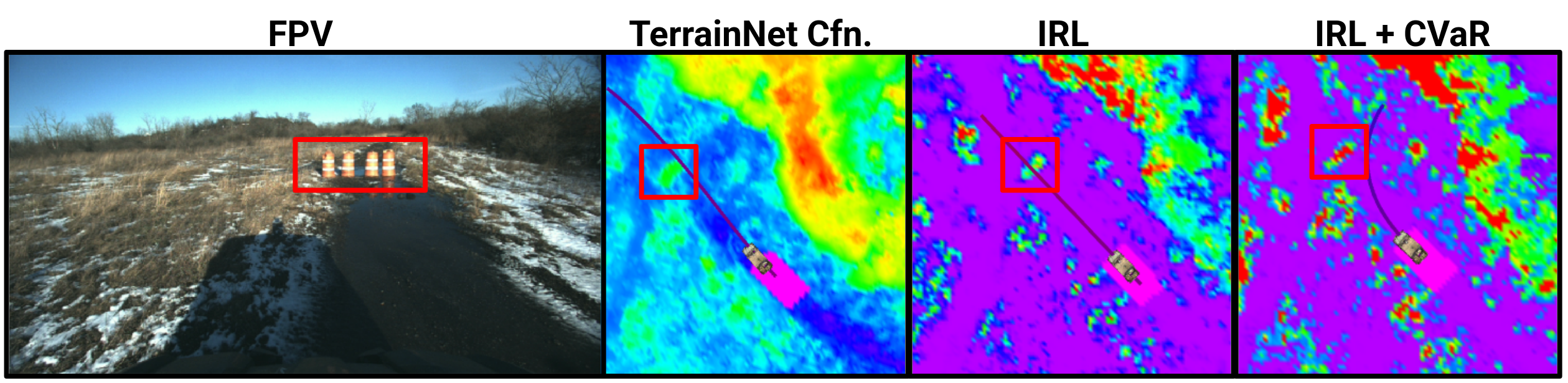}
    \caption{Qualitative figure from the navigation trial. The robot was tasked with navigating around the set of cones in the trail. (Left) Inpaint + IRL without CVaR plans through the cones. (Center) The TerrainNet cfn does not sufficiently distinguish between the cones and the tall vegetation to its side. (Right) IRL with CVaR assigns enough cost to the cones for the vehicle to plan around into the grass.}
    \label{fig:nav_trial_3_qual}
\end{figure}

High-level metrics from each of the autonomy runs are presented in Table \ref{tab:hardware_metrics}. A GPS plot of the interventions with FPV views is presented in Figure \ref{fig:nav_cousrse_gps}. Overall, we observed that the best-performing method in terms of intervention metrics is \textbf{Parallel Trav. + CVaR}. Notably, this was the only method that was able to successfully avoid all of the man-made obstacles, as \textbf{Parallel Trav.} and \textbf{TerrainNet Cfn.} each had a collision, suggesting that adding risk-awareness results in safer navigation. A comparison of the various methods navigating around some man-made cones can be seen in Figure \ref{fig:nav_trial_3_qual}. This scenario required the vehicle to divert from the trail to avoid the cones. Since cones were out-of-distribution from the train set, \textbf{Parallel Trav.} was overly optimistic. Similarly, the semantic inpainting struggled to inpaint the cones in the BEV map, leading \textbf{TerrainNet Cfn.} to understimate the cost. However, using CVaR to capture the semantic uncertainty enabled the robot to avoid the cones. Additionally, we can see that the \textbf{FCN no Inpaint} method was both the slowest and had the largest number of interventions, suggesting that the inpainting had a positive impact on both traversal speed and number of required backtracks. A qualitative example of the value of inpainting is presented in Figure \ref{fig:nav_trial_1_qual} in which the \textbf{FCN no Inpaint} method assigned low cost to an area containing vegetation that was out of direct view. Since the goal point was directly beyond this region, the robot mistakenly started planning into this vegetation, resulting in it getting stuck in a cul-de-sac. However, with inpainting, all other models were able to pick up on the surrounding geometric and semantic cues and inferred that the unobserved area was likely high cost, leading the planner to take a longer but safer path on the trail. We also observed that all methods with inpainting had higher average autonomous speed. This is likely because the inpainting objective resulted in more temporally stable speed predictions. Fluctuations in the speedmap, like those seen in Figure \ref{fig:offline_qual} force the ATV to reduce its speed not just for the offending cell, but also for prior cells as braking cannot occur instantaneously. Unsurprisingly, adding CVaR (thus reducing risk-tolerance) decreased autonomous speed as the robot had to avoid more objects, but reduced the number of interventions.

% \begin{figure}
%     \centering
%     \includegraphics[width=\linewidth]{figs/intervention_fig.png}
%     \caption{GPS plot of the interventions incurred by each method, as well as the general expected path of the navigation trial. Qualitative points of interest are also annotated.}
%     \label{fig:gps_interventions}
% \end{figure}

\begin{table}[]
    \centering
    \begin{tabular}{c||c|c|c|c}
        Method & Col. & Int. & Avg. Speed & Top Speed\\
        \hline
        Parallel Trav. & 2 & 2 & 4.79m/s & 7.66m/s \\
        Terrainnet Cfn. & 1 & \textbf{1} & 5.23m/s & 7.72m/s\\
        FCN no Inpaint & 1 & 4 & 4.33m/s & 6.84m/s\\
        Parallel Trav. + CVaR & \textbf{0} & \textbf{1} & 4.49m/s & 7.16m/s\\
    \end{tabular}
    \caption{Hardware Metrics}
    \label{tab:hardware_metrics}
    \vspace{-6mm}
\end{table}

% \begin{figure*}
%     \centering
%     \includegraphics[width=\linewidth]{figs/nav_trial_pt2.png}
%     \caption{Qualitative figure for point 2 in the navigation trial. The next waypoint is about $50m$ forward-left of the vehicle. The ATV mistakenly navigated right around some cones instead of left. This make it kinematically impossible to drive back to the trail without requiring a backup maneuver.}
%     \label{fig:nav_trial_2_qual}
% \end{figure*}

% \begin{figure}
%     \centering
%     \includegraphics[width=\linewidth]{figs/nav_trial_pt4.png}
%     \caption{Qualitative figure from point 4 in the navigation trial. Dynamically lowering CVaR enables the robot to drive through a corridor of dense vegetation.}
%     \label{fig:nav_trial4_qual}
% \end{figure}

\section{Conclusion} 
\label{sec:conclusion}

In this work, we presented TravSUITE, a fully self-supervised traversability and map inpainting learning framework. We demonstrated that inpainting and risk estimation play an important role in learned traversability estimation methods in both offline and hardware results. Furthermore, we have demonstrated a straightforward technique for generating large-scale geometric-semantic inpainting supervision that requires no additional engineering overhead. We believe that this is a significant step forward for self-supervised traversability estimation methods in unstructured terrain.

While TravSUITE represents step forward for learning-based local traversability, there are some limitations of TravSUITE worth highlighting. First, while this method is environment and robot-generic in principle, additional experiments in additional biomes and platforms are needed to fully substantiate this claim. Second, while we provide a rudimentary uncertainty estimation technique that seems to work well in practice, a more principled treatment of uncertainty (such as \cite{jung2025uncertainty, cai2024pietra}) may improve performance in OOD. Lastly, TravSUITE only performs local traversability estimation and cannot reason about information beyond the mapping horizon.

Future work will include exploring additional sources of high-quality self-supervised data and demonstrating our system in more environments such as multi-floor buildings and on other platforms such as legged robots. We are also interested in co-training additional tasks such as full 3d path planning and segmentation that necessitate a volumetric representation. We are also interested in exploring non-metric representations to enable long-range traversability estimation and planning \cite{fahnestock2025far, schmittle2025long}. 

\section*{Acknowledgments}

The authors would like to thank Matthew Sivaprakasam, Micah Nye and Parv Maheshwari for their help with field testing and feedback in the paper-writing stage. This work was supported by DSO National Laboratories (DSO) Contract \#DSOCO25020. This work used Bridges-2 at PSC through allocation cis220039p from the Advanced Cyberinfrastructure Coordination Ecosystem: Services \& Support (ACCESS) program which is supported by NSF grants \#2138259, \#2138286, \#2138307, \#2137603, and \#213296. This work was supported in part by Field AI, Inc.

%% Use plainnat to work nicely with natbib. 

\bibliographystyle{plainnat}
\bibliography{references}

\newpage

% \documentclass[conference]{IEEEtran}
% \usepackage{times}

% % numbers option provides compact numerical references in the text. 
% \usepackage[numbers]{natbib}
% \usepackage{multicol}
% \usepackage[bookmarks=true]{hyperref}

% \usepackage{bbm}
% \usepackage{amsmath}
% \usepackage{amssymb}
% \usepackage{graphicx}
% \usepackage[ruled]{algorithm2e}

% \DeclareMathOperator*{\argmax}{arg\,max}
% \DeclareMathOperator*{\argmin}{arg\,min}

% \pdfinfo{
%    /Author (Homer Simpson)
%    /Title  (Robots: Our new overlords)
%    /CreationDate (D:20101201120000)
%    /Subject (Robots)
%    /Keywords (Robots;Overlords)
% }

% \begin{document}

\appendix

% % paper title
% \title{Appendix for TravSUITE}

% % You will get a Paper-ID when submitting a pdf file to the conference system
% \author{\authorblockN{Samuel Triest\authorrefmark{1},
% Amirreza Shaban\authorrefmark{2},
% David D. Fan\authorrefmark{2}, 
% Wenshan Wang\authorrefmark{1} and
% Sebastian Scherer\authorrefmark{1}}
% \authorblockA{\authorrefmark{1}Robotics Institute, Carnegie Mellon University}
% \authorblockA{\authorrefmark{2}Field AI\\ Email: striest@andrew.cmu.edu}
% }

% \maketitle

\section{Additional Experimental Analysis}

\begin{figure}
    \centering
    \includegraphics[width=\linewidth]{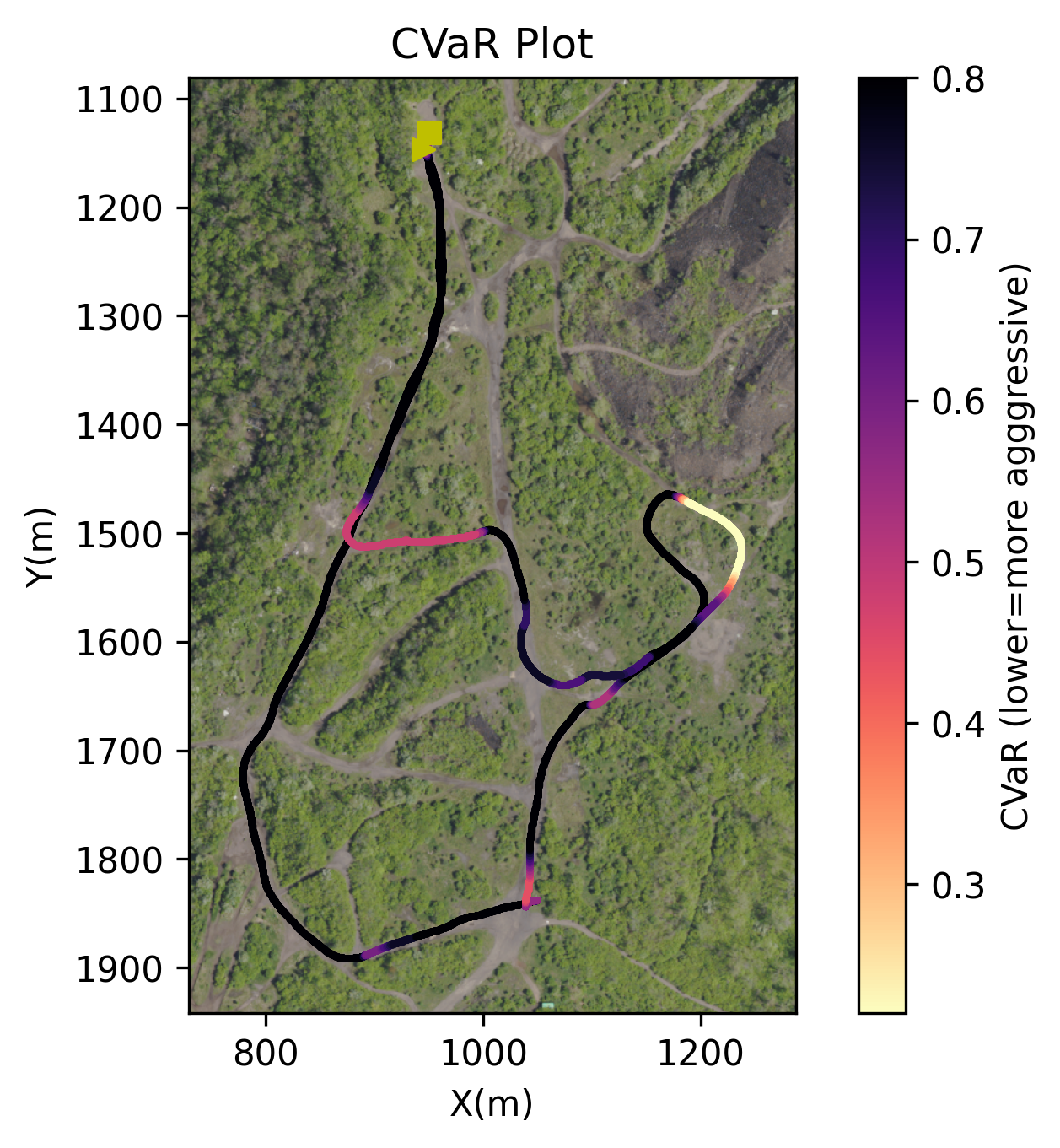}
    \caption{Plot of the CVaR value during the \textbf{Parallel Trav. + CVaR} autonomy run.}
    \label{fig:cvar_gps}
\end{figure}

In figure \ref{fig:cvar_gps}, we provide a GPS plot of the online CVaR value during the \textbf{Parallel Trav. + CVaR} experiment. We can see that maximum risk aversion ($\nu=0.8$) is sufficient for the majority of the course, with a few more aggressive areas ($\nu=0.2-0.5$) being required for a few areas with denser vegetation.

\begin{figure*}
    \centering
    \includegraphics[width=\linewidth]{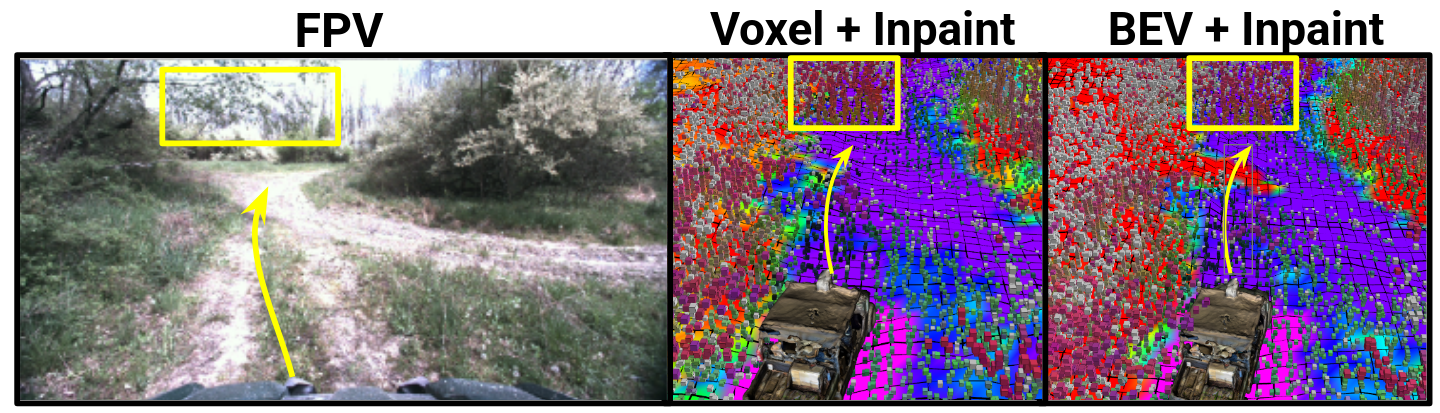}
    \caption{Qualitative example comparing voxel inputs versus BEV. (Yellow box): There is a low-hanging canopy over the trail, which the demonstration trajectory (yellow arrow) goes under. The voxel network is correctly able to predict the trail as low cost, while the BEV network misclassifies the trail as an obstacle due to the limited vertical representativeness of the BEV representation.}
    \label{fig:voxel_qual}
\end{figure*}

We also present an additional qualitative example of the robot navigating under a significant overhang (\textbf{Test 2}) in Figure \ref{fig:voxel_qual} in which we compare two of our networks trained on the full traversability and inpainting objective, where the only difference is whether the input representation was in 3D or BEV (Voxel and BEV from Table 4 in the main paper). Here, we can see that the voxel-based network correctly discards the information from the overhanging canopy and predicts low cost, while the BEV-based network mis-classifies the trail under the canopy as an obstacle.

\section{Voxel Mapping Implementation Details}

We will describe in greater detail the procedure we use for voxel mapping. We implement our voxel mapper entirely from torch \cite{paszke2017automatic} and torch\_scatter \cite{pytorch_scatter} operations, enabling efficient parallel computation on GPUs. For the purposes of this section, we will assume that the vehicle receives a state estimate $x$, point cloud $P$ and image $I$ at every timestep. We will assume that all sensor streams are synchronized, though in practice we perform motion compensation to temporally align the image and point cloud.

\subsection{Semantic Feature Extraction and Lifting}

First, in order to aggregate our measurements in 3D space, we first extract meaningful semantic content from our image stream $I_t$, and lift it using the pointcloud $P_t$. We use two choices of semantic features which are common in prior work.

\begin{enumerate}
    \item \textbf{VFM semantics}: A common approach to extracting semantics is to perform open-set segmentation on VFM features. In line with this, we leverage Talk2Dino \cite{barsellotti2025talking} with text prompts from the Rellis3d semantic ontology \cite{jiang2020rellis3d} (with an additional snow class), and map the similarity between each pixel feature $z_I$ and text embedding $z_{c_i}$. Empirically, we found that ``sharpening'' the cosine similarities (by multiplying by a fixed $\beta$ before softmax) produced better segmentation results (Eq. \ref{eq:seg_csim}).
    \item \textbf{Raw VFM}: Another popular approach to feature extraction is to simply map the VFM features directly \cite{triest2024velociraptor, alama2025rayfronts}. In this case, we choose to use the visual encoder used in RayFronts \cite{alama2025rayfronts} (Radiov3 \cite{ranzinger2023radio} with Siglip2 adaptor and NAClip attention). We then perform a principal component analysis to reduce the feature dimension further to the top 32 components.
\end{enumerate}

\begin{multline}
    \label{eq:seg_csim}
    p(c_i) = softmax(\beta D_C)_i, \\
    \quad D_C = [\hbox{Talk2DINO}(z_I, z_{c_0}), \dots \hbox{Talk2DINO}(z_I, z_{c_N})]
\end{multline}

In either case, the end result is a ``feature image'' $I^f_t \in \mathbb{R}^{W//k \times H//k \times n_f}$, where $k$ is the reduction in resolution from the original image. 

Given the feature image, we must then lift our FPV image features into 3D. Fortunately, this process is straightforward via projection of points $[x,y,z] \in P_t$ into the image frame using Eq. \ref{eq:point_projection}, given camera intrinsics $\mathbf{K}$ and camera-lidar extrinsics $\mathbf{R}, \mathbf{t}$. (Note that $\mathbf{S} \triangleq \hbox{diag}([\frac{1}{16}, \frac{1}{16}, 1])$ is an additional scale matrix used to account for lower resolution of the feature image).

\begin{equation}
    \label{eq:point_projection}
    [u, v, w]^T = \mathbf{K} \mathbf{S} [\mathbf{R} | \mathbf{t}] [x, y, z, 1]^T, \quad i = \frac{u}{w}, \quad j = \frac{v}{w}
\end{equation}

We apply several known tricks to improve the sharpness and accuracy of our feature pointcloud, such as bilinear interpolation and per-pixel depth binning \cite{premebida2016high, alama2025rayfronts}. While this reduces the overall number of points that receive a VFM feature, it is essential to maintain sufficient sharpness of features, especially near edges of objects.

\subsection{Voxel Local Mapping with Features}

Given our streams of feature point clouds $P_t$, we can now accumulate them into a voxel grid. This voxel grid $V = <\mathcal{M}, \{i, f_i\}>$ can be expressed as a local spatial extent (containing an origin, length and resolution $\mathcal{M}_o, \mathcal{M}_l, \mathcal{M}_r$, each in $\mathbb{R}^3$), a set of voxels which can each be represented as a unique integer index $\{i \in [0, N_x N_y N_z)\}$ and feature vector $\{f_i \in \mathbb{R}^{n_f}\}$ (where $N_{x, y, z} = \lfloor \frac{\mathcal{M}^{x, y, z}_l} {\mathcal{M}^{x, y, z}_r} \rfloor$). In practice the voxels are stacked into one and two dimensional torch \cite{paszke2017automatic} arrays, respectively. The sparsity of this representation enables the voxel grid to be sufficiently efficient and compact to be suitable for real-time, online computation. Furthermore, this representation is analogous to that used by common sparse convolution libraries such as spconv \cite{spconv2022}. The remainder of this section will detail the implementation of our voxel mapper, given our voxel representation.

The goal of the voxel mapper is to maintain and aggregate streams of feature pointclouds (transformed into an inertial frame via odometry) over time. Given the pose and feature pointcloud at the current timestep, and voxel grid at the previous timestep (and a fixed local extent $\mathcal{M}_{local}$), we can perform voxel mapping of features by performing the following operations:

\begin{enumerate}
    \item EXTENT($x_t, \mathcal{M}_{local}$): Compute the current extent $\mathcal{M}_t$ from the current pose and local extent.
    \item DISCRETIZE($x_t, \mathcal{P}_t, \mathcal{M}_t$): Compute a voxel grid $\dot{V}$ from the current sensor measurements.
    \item SHIFT($V_{t-1}, \mathcal{M}_t$): Shift the previous voxel grid into the current frame ($V'_{t-1}$). 
    \item MERGE($V'_{t-1}, \dot{V}$): Update the shifted voxel grid with the instantaneous voxel grid.
\end{enumerate}

Fortunately, efficient instantiations of the above methods can be implemented using a combination of torch \cite{paszke2017automatic} and torch\_scatter \cite{pytorch_scatter} operations. 

\subsubsection{EXTENT}
    The EXTENT operation can be implemented via Equation \ref{eq:EXTENT}. Note that we choose to have the extent origin be a multiple of the resolution to simplify the implementation of the downstream operations. 
    
    \begin{equation}
        \label{eq:EXTENT}
        EXTENT(x_t, \mathcal{M}_{local}) = \begin{bmatrix}
            o_t = \lfloor \frac{x_t}{r_{local}} \rfloor  r_{local} \\
            l_t = l_{local} \\
            r_t = r_{local} \\
        \end{bmatrix}
    \end{equation}
    
\subsubsection{DISCRETIZE}
    The core of the DISCRETIZE operation relies on a mapping between points in the mapping extent and unique integer indices defined in Equation \ref{eq:raster}. The instantaneous voxel grid can then be computed by computing the set of unique indices $i$, and defining how to compute a feature from the set of points with the same index $i$. We choose to have the voxel take the feature of the point closest to the robot pose, though other aggregation functions such as average are reasonable as well.

    \begin{equation}
        \label{eq:raster}
        i = raster_M(x,y,z) = \begin{bmatrix}
            \lfloor \frac{x - o_x}{r_x} \rfloor \\
            \lfloor \frac{y - o_y}{r_y} \rfloor \\
            \lfloor \frac{z - o_z}{r_z} \rfloor \\
        \end{bmatrix}^T
        \begin{bmatrix}
            N_y N_z \\
            N_z \\
            1 \\
        \end{bmatrix}
    \end{equation}

\subsubsection{SHIFT}
    Given our choice of index assignment in Equation \ref{eq:raster}, we can efficiently perform transforms of integer multiples of the voxel map resolution via Equation \ref{eq:shift}.

    \begin{equation}
    \label{eq:shift}
        i' = i + raster_\mathcal{M}(dX), \quad dX = \lfloor \frac{o-o'}{r} \rfloor
    \end{equation}

    Indices (and corresponding features) that fall outside of the mapping volume are discarded.
    
\subsubsection{MERGE}
    Finally, in order to merge two voxel grids together, the resulting set of indices can be computed as the set intersection of the indices in $V'_{t-1}$ and $\dot{V}$. If the index is in only one set, the corresponding feature is copied over. Otherwise, the features are combined via an exponential moving average.

    \begin{flalign}
        \label{eq:feat_ema}
        MERGE(V_1, V_2) = <I, F>, \hbox{where} \\
        I = I_1 \cup I_2 \\
        F = \begin{cases}
            f^1_i & i \notin I^1 \\
            f^2_i & i \notin I^2 \\
            \alpha f^1_i + (1 - \alpha) f^2_i & \hbox{otherwise}
        \end{cases}
    \end{flalign}

    We perform an additional, optional raycasting operation using $P_t$ to help with noisy observations and dynamic objects. For every voxel, we maintain an additional hit and miss feature. Voxels are initialized with $hits=1, misses=0$, and the counts are summed in the MERGE operation. A voxel's miss counter is incremented if a ray from $x_t$ to any point in $P_t$ passes through the voxel. At every timestep, voxels with a passthrough ratio ($\frac{misses}{hits+misses}$) over 0.75 are discarded.

\subsection{Runtime Experiments}

We ran an experiment to compare the speed of our voxel mapper to an elevation mapping-based alternative. In both cases, the local mapper was tasked to map the environment as well as the VFM features from Radiov3 \cite{ranzinger2023radio}. In Table \ref{tab:timing_stats}, we report the mean, minimum and maximum times (in $ms$) over a run of roughly 2000 frames for each subcomponent of the elevation and voxel pipelines. We observe that the total time per frame between each mapping pipeline is very similar, totaling a speed difference of fewer than $2ms$ on average per timestep. We note that while the voxel mapping is slower than the elevation mapping, the fraction of time spent per cycle on the mapping step is a very small fraction of the total mapping speed. Visual and BEV feature extraction, which includes computation of image features and terrain estimation, accounts for $70-80\%$ of the total mapping pipeline runtime for both approaches. In either case, both methods are capable of running at around $8-10hz$ (notwithstanding the ROS2 interface). We note that the voxel-based network, however, is considerably slower, and may not scale as effectively to larger map sizes.

\begin{table*}[]
    \centering
    \begin{tabular}{c||c|c|c|c|c}
              & Image Feature Extraction & Feature Lifting & Mapping & BEV Feature & Inference \\
        \hline
        Voxel & 26.3/25.6/27.0 & 2.8/2.6/3.9 & 8.5/8.1/11.2 & 12.3/12.0/13.6$^*$ & 31.1/74.7/103.5 \\
        BEV   & 26.4/25.7/27.9 & 2.8/2.7/4.9 & 3.5/3.2/7.3  & 15.8/14.6/19.4 & 25.9/27.2/32.0  \\
    \end{tabular}
    \caption{Timing speeds (mean/min/max in ms)}
    \label{tab:timing_stats}
\end{table*}

\section{Additional Training Details}

We provide additional implementation details that were omitted from the main paper.

\subsection{Computing Behavior Labels}

Computation of the behavior labels for $C_t, S_t$ require computation of BEV-space state visitation distributions and speed supervision. These must both be of the same width and weight of the behavior map $\Pi_t$. In order to generate these labels, we must have access to 1) the current robot state $x$, which must have a planar position $x,y = x_p$ and orientation $x_\theta$, 2) the mapping extent $\mathcal{M}$, and 3) optionally a footprint, defined as a set of positional offsets $\Delta$ to account for the robot occupying multiple BEV cells. Given this, we are able to generate labels by iterating through every state $x$ in the trajectory $\tau$, optionally expanding it into multiple states $x'$ via the footprint, computing the map index of every cell, and scattering the appropriate quantity into a $N_x \times N_y$ buffer. For state visitations, we scatter-sum ones for every state and normalize the buffer to sum to one. For speeds, we scatter-average the corresponding speed for every point $x'$.

% im scared this is too close to the vrap supplemental
\begin{algorithm}
\DontPrintSemicolon
\caption{Computation of $SV(\tau)$ and $\hat{S}(\tau)$}
\label{algo:label_computation}
\KwIn{\\
        \quad Trajectory $\tau$, \\
        \quad Footprint $F = \{\Delta_0 \hdots \Delta_n\}$ , \\
        \quad Map Metadata $o_x, o_y, r_x, r_y, n_x, n_y$
}
\KwOut{\\
    \quad State Visitation Distribution $\mathbf{D} \in \mathbb{R}^{n_x \times n_y}$ \\
    \quad Speed Label Map $\hat{\mathbf{S}} \in \mathbb{R}^{n_x \times n_y}$ \\
}

$l_{x,y} = n_{x, y} r_{x, y}$ \;
$\mathbf{D}_{i, j} = 0, \quad \forall i,j \in \big [ [0 \hdots n_x] \times [0 \hdots n_y] \big ]$ \;
$\mathbf{\hat{S}}_{i, j} = 0, \quad \forall i,j \in \big [ [0 \hdots n_x] \times [0 \hdots n_y] \big ]$ \;

\For{$x \in \tau$}
{
    $s = ||x_v||_2$ \;
    $R = rotm(x_q)$ \;
    \For{$\Delta \in F$}
    {
        $x'_x, x'_y, x'_z = x_p + R\Delta$ \;
        \If{$x'_x \in [o_x, o_x+l_x)$ and $x'_y \in [o_y, o_y+l_y)$}
        {
            $i = \lfloor \frac{x'_x - o_x}{r_x} \rfloor, j = \lfloor \frac{x'_y - o_y}{r_y} \rfloor$ \;
            $\mathbf{D}_{i, j} = \mathbf{D}_{i, j} + 1$ \;
            $\hat{\mathbf{S}}_{i, j} = \hat{\mathbf{S}}_{i, j} + s$ \;
        }
    }
}

$\hat{\mathbf{S}} = \frac{\hat{\mathbf{S}}}{\mathbf{D}}$ \;

$\mathbf{D} = \frac{\mathbf{D}}{\sum \mathbf{D}}$ \;

return $\mathbf{D}, \hat{\mathbf{S}}$ \;
\end{algorithm}

\subsection{Estimating the Terrain Surface}

In order to properly compute BEV features from voxel grids, it is important to have an understanding of the height of the terrain surface for every cell in the BEV map. However, due to occlusions and noise, it is unreliable to simply assume that the minimum-height voxel in a column belongs to the terrain surface. As such, we follow the approach used in prior work \cite{wellington2005interacting, maturana2022semantic} and use a Markov random field to estimate the underlying terrain surface for our BEV features. We then use the estimated terrain surface to determine the diff feature and overhang threshold for every cell.

\subsection{Computing Hindsight Labels}

Since are able to derive BEV labels $B^*$ from any voxel map $V^*$, our hindsight labeling algorithm focuses on $V^*_t$, as $B^*_t$ can be obtained by running BEV feature extraction on $V^*_t$.

Prior work typically relies on producing $V^*_t$ by aggregating observations in a fixed time window, e.g. [$t-k, t+k$]. While this approach is straightforward, it can lead to suboptimal supervision if the speed of demonstrations varies significantly. For instance, if the robot is traveling very quickly, only a small number of frames in the window [$t-k, t+k$] will actually fall within the the mapping volume at time $t$. Conversely, if the robot is stationary, then all observations in [$t-k, t+k$] wil be the same and $V^*_t$ will degenerate to $V_t$.

To address this, we propose a simple two-pass algorithm for computing voxel supervision that is invariant to robot speed. For ease of presentation, we define two functions $V_t = \hbox{voxelmap}(V_{t-1}, x_t, O_t, \mathcal{M}_{local})$, which corresponds to running the voxel mapper on the current sensing data with a given local extent, and $\hbox{crop}(V_t, \mathcal{M})$, which extracts the voxels from $V_t$ that lie within the mapping extent $\mathcal{M}$ (note that the cropping extent will differ from the extent of $V_t$).

\begin{figure}
    \centering
    \includegraphics[width=\linewidth]{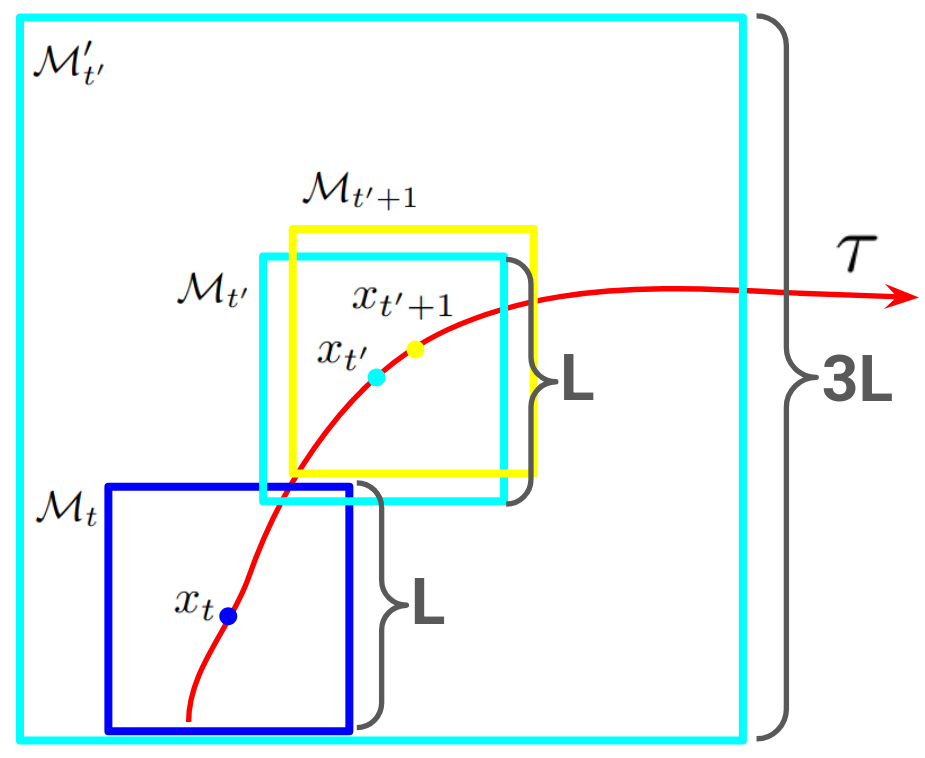}
    \caption{A simple 2d example to highlight the intuition behind the voxel hindsight generation algorithm. (Blue) The state $x_t$ and corresponding mapping volume $\mathcal{M}_t$. (Cyan) The state $x_{t'}$ and corresponding mapping volume $\mathcal{M}_{t'}$. Note that $\mathcal{M}_{t'}$ intersects with $\mathcal{M}_t$ and $\mathcal{M}_{t'+1}$ does not. Since $\mathcal{M}_{t'}$ intersects with $\mathcal{M}_t$, it must be the case that a volume with triple the side length, centered at $x_t'$ (i.e. $\mathcal{M}'_{t'}$) must fully contain $\mathcal{M}_t$.}
    \label{fig:voxel_hindsight_example}
\end{figure}

At a high level, this algorithm works by computing a target time $t'$ for all times $t$ in the trajectory, where $t' \geq t$ is the last time at which the future mapping extent $\mathcal{M}_{t'}$ intersects the current mapping extent $\mathcal{M}_{t}$. This timestep is significant because the voxel grid $V_{t'+1}$ will share no voxels with $V_t$. Thus, $V^*_t$ only contains contributions from voxel grids $V_0$ to $V_{t'}$ (note that this discounts backtracks, however accounting for backtracks of arbitrary length is challenging because it requires perfect state estimation). The intuition for this is presented in Figure \ref{fig:voxel_hindsight_example}. Since this time interval is contiguous and monotonically increasing, we can re-purpose the original voxel mapper to compute $V^*_t$ by deriving it from $V_{t'}$ . However, since $x_t$ and $x_{t'}$ are different states, we must use a larger mapping extent such that the new extent $\mathcal{M}'_{t'}$ at the target time $t'$ fully contains the original extent $\mathcal{M}_{t}$ at the current time $t$. Fortunately, since we know that $\mathcal{M}_{t'}$ intersects $\mathcal{M}_{t}$, it must be the case that a volume with triple the side length, centered at $x_{t'}$, fully contains $\mathcal{M}_{t}$. Given this, we can get $V^*_t$ by cropping $V'_{t'}$ with extent $\mathcal{M}_t$. The detailed algorithm is presented in Algorithm \ref{algo:label_computation}. It is important to note that while this algorithm has $O(T^2)$ volume intersection operations, it maintains $O(T)$ voxel processing operations. This is the same asymptotic complexity as the methods used by prior work \cite{shaban2022semantic, meng2023terrainnet} with respect to voxel mapping operations, which comprise nearly all of the total computation time. Once we obtain $V^*_t$ for every timestep in a trajectory, we then generate $B^*_t$ by running the BEV feature extractor on $V^*_t$.

\begin{algorithm*}
\DontPrintSemicolon
\caption{Voxel Hindsight Label Generation}
\label{algo:label_computation}
\KwIn{\\
        \quad Trajectory $\tau = x_{0:t}$, \\
        \quad Observations $O_{0:T}$ , \\
        \quad Map Extent $\mathcal{M}_{local}$ \\
}
\KwOut{\\
    \quad $V^*_t, \quad \forall t$ \\
}

\tcp{First Pass}

$\mathbf{T}^{end} = [t, \quad \forall t \in [0 \hdots T]]$ \hfill $\triangleleft$ initialize end time for all timesteps \; 

\For{$t_1 \in [0 \hdots T]$}{
    \For{$t_2 \in [t_1 \hdots T]$}{
        $\mathcal{M}_1 = x_{t_1} + \mathcal{M}_{local}$ \hfill $\triangleleft$ compute mapping volume at $t_1$ \;
        $\mathcal{M}_2 = x_{t_2} + \mathcal{M}_{local}$ \hfill $\triangleleft$ compute mapping volume at $t_2$ \;
        \If{$\mathcal{M}_1 \cap \mathcal{M}_2 \neq \emptyset$}{
            $\mathbf{T}^{end}_{t_1} := min(T, t_2+1)$ \hfill $\triangleleft$ if the mapping volumes intersect, increase the end time for $t_1$\;
        }
    }
}

\tcp{Second Pass}
$\mathbf{V}^* = [\emptyset, \quad \forall t \in [0 \hdots T]]$ \hfill $\triangleleft$ initialize output maps for all timesteps \;

$\mathcal{M}'_{local} = (\mathcal{M}^o_{local} - \mathcal{M}^l_{local}, \mathcal{M}^r_{local}, 3 \mathcal{M}^l_{local})$ \hfill $\triangleleft$ initialize new mapping extent with 3x size \;

$V_{-1} = \emptyset$ \hfill $\triangleleft$ initialize empty voxel map \;

\For{$t \in [0 \hdots T]$}{
    $V_t = \hbox{voxelmap}(V_{t-1}, x_t, O_t, \mathcal{M}'_{local})$ \hfill $\triangleleft$ voxel map the current data with larger extent\;
    \For{$[t' \quad s.t. \quad t' = \mathbf{T}_t]$}{
        $\mathcal{M}_{t'} = x_{t'} + \mathcal{M}_{local}$ \hfill $\triangleleft$ compute target mapping volume for $t'$ \;
        $\mathbf{V}^*_{t'} = \hbox{crop}(V_t, \mathcal{M}_{t'})$ \hfill $\triangleleft$ crop out voxels in the target volume \;
    }
}

\end{algorithm*}

\subsection{Hyperparameters}

Provided in this section are the various hyperparameters for our method (Tables \ref{tab:local_mapping_params} \ref{tab:net_params} \ref{tab:training_params} \ref{tab:hardware_params}). For all experiments, the voxel and BEV metadata match in the x and y dimensions.

\begin{table}[]
    \centering
    \begin{tabular}{c|c}
        Parameter & Value \\
        \hline
        Mapping Origin ($\mathcal{M}_o$) & $-50m \times -50m \times-10m$ \\
        Mapping Length ($\mathcal{M}_l$) & $100m \times 100m \times 20m$ \\
        Mapping Resolution ($\mathcal{M}_r$) & $0.4m \times 0.4m \times 0.1m$ \\
        Mapping Cell Counts ($\mathcal{M}_n$) & $250 \times 250 \times 200$ \\
        Input Image Shape & $848 \times 448 \times 3$ \\
    \end{tabular}
    \caption{Local Mapping Parameters}
    \label{tab:local_mapping_params}
\end{table}

\begin{table}[]
    \centering
    \begin{tabular}{c|c}
        Parameter & Value \\
        \hline
        Voxel Encoder & 3D ResNet Block $\times 4$ \\
        BEV Encoder & ResNet Block $\times 4$ \\
        Backbone & UNet Block $\times 3$ \\
        VFM Decoder & ResNet Block $\times 2$ \\
        Seg. Decoder & ResNet Block $\times 2$ \\
        Geom. Decoder & ResNet Block $\times 2$ \\
        Cost Decoder & ResNet Block $\times 2$ \\
        Speed Decoder & ResNet Block $\times 2$ \\
        Cost Range & $[0, 5]$ \\
        Cost N Bins & 32 \\
        Speed Range & $[0, 15]m/s$ \\
        Speed N Bins & 32 \\
    \end{tabular}
    \caption{Network Parameters}
    \label{tab:net_params}
\end{table}

\begin{table}[]
    \centering
    \begin{tabular}{c|c}
        Parameter & Value \\
        \hline
        Goal Weight ($K_2$, Eq. 3) & 5.0 \\
        Vehicle Footprint & $4m \times 2m$ \\
        Behavior Weight ($\lambda_1$, Eq. 4) & 1.0 \\
        Inpaint Weight ($\lambda_2$, Eq. 4) & 1.0 \\
        Geom Inpaint Weight ($\lambda_{geom}$, Eq. 9) & 1.0 \\
        VFM Inpaint Weight ($\lambda_{vfm}$, Eq. 9) & 1.0 \\
        Seg Inpaint Weight ($\lambda_{seg}$, Eq. 9) & 1.0 \\
        Speed Contrastive Weight ($\lambda_s$, Eq. 6) & 0.05 \\
        Optimizer & Adam \cite{kingma2014adam}
    \end{tabular}
    \caption{Training Parameters}
    \label{tab:training_params}
\end{table}

\begin{table}[]
    \centering
    \begin{tabular}{c|c}
        Parameter & Value \\
        \hline
        CVaR min/max & $[0.0, 0.8]$ \\
        CVaR vel min/max ($v_{low}, v_{high}$, Eq. 8) & $[2, 5]m/s$ \\
        CVaR step ($\epsilon$, Eq. 8) & 0.02 \\
        Lat Acc. Max & $5m/s^2$\\
    \end{tabular}
    \caption{Hardware Experiment Parameters}
    \label{tab:hardware_params}
\end{table}

% \bibliography{references}

% \end{document}

\end{document}